\documentclass[10pt,twocolumn,letterpaper]{article}

\usepackage[pagenumbers]{cvpr} 

\usepackage{graphicx}
\usepackage{amsmath}
\usepackage{amssymb}
\usepackage{booktabs}
\usepackage{multirow}
\usepackage{epsfig}
\usepackage[accsupp]{axessibility}  
\usepackage{caption}
\usepackage[pagebackref,breaklinks,colorlinks]{hyperref}

\usepackage[capitalize]{cleveref}
\crefname{section}{Sec.}{Secs.}
\Crefname{section}{Section}{Sections}
\Crefname{table}{Table}{Tables}
\crefname{table}{Tab.}{Tabs.}

\def\cvprPaperID{11053} 
\def\confName{CVPR}
\def\confYear{2022}

\begin{document}

\title{Neural Emotion Director: Speech-preserving semantic control of facial expressions in “in-the-wild” videos}

\author{
Foivos~Paraperas~Papantoniou$^1$ \quad
Panagiotis~P.~Filntisis$^1$    \quad
Petros~Maragos$^1$   \quad
Anastasios~Roussos$^{2,3}$ \\
\small{$^{1}$School of Electrical \& Computer Engineering, National Technical University of Athens, Greece}
\\
\small{$^{2}$Institute of Computer Science, Foundation for Research \& Technology - Hellas (FORTH), Greece}
\\
\small{$^{3}$College of Engineering, Mathematics and Physical Sciences, University of Exeter, UK}
}


\twocolumn[{%
\renewcommand\twocolumn[1][]{#1}%
\maketitle
\begin{center}
\vspace{-0.3cm}
    \centering
    \captionsetup{type=figure}
    \includegraphics[trim=0 0 0 0, clip, width=.9\textwidth]{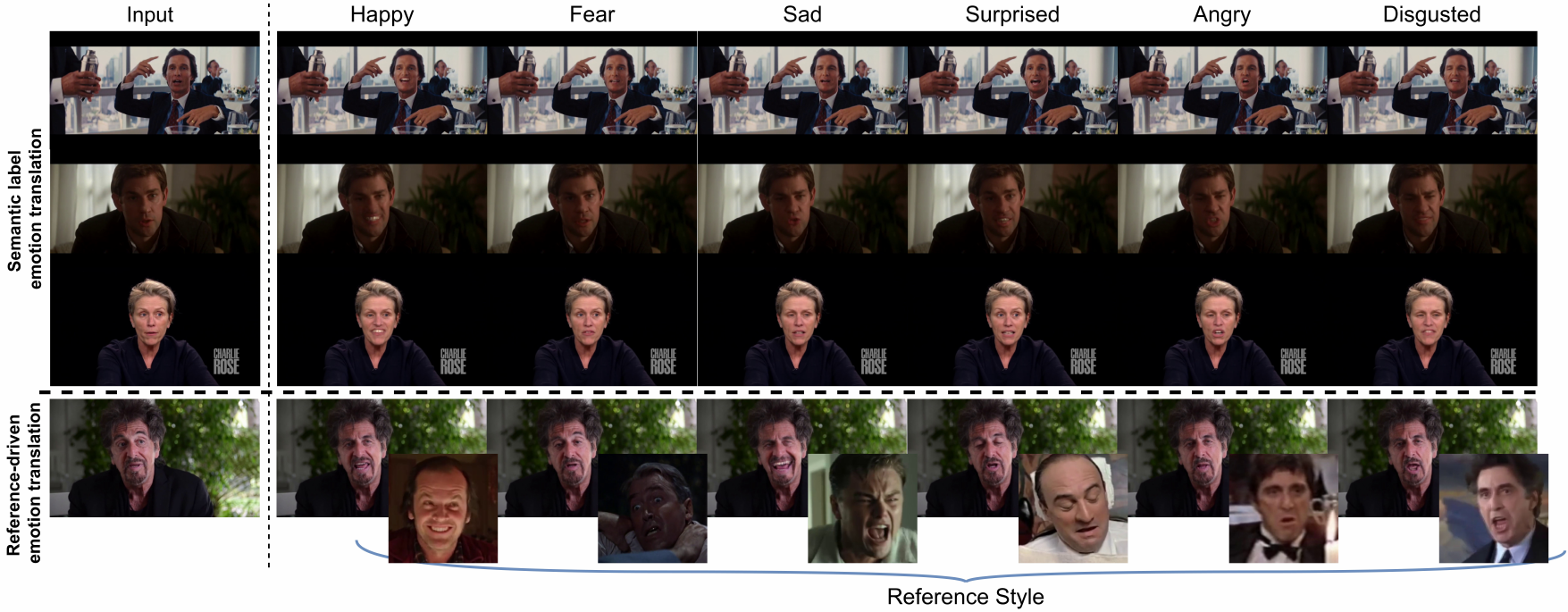}
    \vspace{-0.3cm}
    \captionof{figure}{Our \textit{Neural Emotion Director (NED)} can manipulate facial expressions in input videos while preserving speech, conditioned on either the semantic emotional label (top part of figure), or on an external reference style as extracted from a reference video (bottom part). Please zoom in for details. Video results can be found at: \url{https://foivospar.github.io/NED/}}
    \label{fig:ned_teaser}
\end{center}%
}]
\begin{abstract}
In this paper, we introduce a novel deep learning method for photo-realistic manipulation of the emotional state of actors in ``in-the-wild'' videos. The proposed method is based on a parametric 3D face representation of the actor in the input scene that offers a reliable disentanglement of the facial identity from the head pose and facial expressions. It then uses a novel deep domain translation framework that alters the facial expressions in a consistent and plausible manner, taking into account their dynamics. Finally, the altered facial expressions are used to photo-realistically manipulate the facial region in the input scene based on an especially-designed neural face renderer. To the best of our knowledge, our method is the first to be capable of controlling the actor’s facial expressions by even using as a sole input the semantic labels of the manipulated emotions, while at the same time preserving the speech-related lip movements. We conduct extensive qualitative and quantitative evaluations and comparisons, which demonstrate the effectiveness of our approach and the especially promising results that we obtain. Our method opens a plethora of new possibilities for useful applications of neural rendering technologies, ranging from movie post-production and video games to photo-realistic affective avatars.
\end{abstract}

\section{Introduction}
\label{sec:intro}


Photo-realistically manipulating faces in images or videos has received substantial attention lately, with impressive results that expand the range of creative video editing, content creation and the VFX industry. Yet, when it comes to altering the facial emotion in videos, existing techniques exhibit severe limitations. The importance of this type of manipulation is clearly illustrated when shooting movies, as capturing the desired actor's emotion typically requires multiple efforts, despite the uttered words being predefined. A robust solution to emotion editing would conveniently place the manipulation of facial performance in the post-production stage.

In the past, the problem has been tackled by assuming that different recordings of the same script being acted in multiple emotions are available; hence, enabling to switch or blend between takes, after perfect synchronization has been achieved~\cite{FaceDirector}. However, in a more realistic scenario, one would like to make, \eg, a \textit{neutral} actor look \textit{happy}, without using pre-existing footage. Combining performances from unpaired data is much more challenging.

Recently, \textit{image-to-image translation} has been successfully applied to emotion editing by casting the problem in the image space~\cite{StarGAN,sdapolito2021GANmut}. These methods deal with static images, where altering the mouth shape (\eg opening a closed mouth to show surprise) is acceptable, if not desired. However, without placing any specific constraint on the mouth region, lip synchronization may be lost when they are directly applied to video sequences.

More related to this task, is the \textit{face reenactment} problem, where the facial performance of a source actor is transferred to a target one, making the latter mimic the expressions of the former. State-of-the-art techniques~\cite{DVP,head2head++,zakharov_few-shot_2019} achieve compelling photo-realism by training a neural renderer conditioned on a facial representation (\eg 3DMMs). Nonetheless, this is substantially different from semantically controlling the target actor, since the expressions are merely copied from another subject. Instead, we would like to edit the actor's own expressions based on the desired emotion, while preserving the mouth motion. Recent methods address only one aspect of this problem. For instance, DSM~\cite{DSM21} generates novel expressions based on emotional labels without retaining the original speech, while ~\cite{neural_style_preserving_dubbing} preserves mouth movements, but the proposed manipulation is limited to matching a single target speaking style.

In this work, we propose a hybrid method, in which a parametric 3D face representation is translated to different domains, and then used to drive synthesis of the target face by means of a video-based neural renderer. 
Our method, which we call \textit{Neural Emotion Director (NED)}, achieves photo-realistic manipulation of the emotional state of actors in ``in-the-wild'' videos, see \eg~Fig.~\ref{fig:ned_teaser}. 
It can translate a facial performance to any of the 6 basic emotions (angry, happy, surprise, fear, disgust, sadness) plus neutral, using only as input its semantic label, while retaining the original mouth motion. It also allows to attach a specific style 
to the target actor, without requiring person-specific training. This means that the reference style can be extracted at test time from any given video: our system can, for example, make Robert De Niro yell in the way of Al Pacino, without ever seeing footage of the latter during training. Our contributions can be summarized as follows:
\\
$\bullet$ To the best of our knowledge, we propose the first video-based method for ``directing" actors in ``in-the-wild" conditions, by translating their facial expressions to multiple unseen emotions or styles, without altering the uttered speech.
\\
$\bullet$ We introduce an \textit{emotion-translation} network, which we call \textit{3D-based Emotion Manipulator}, that receives a sequence of expression parameters and translates them to a given target domain or a reference style and is trained on non-parallel data. We train this network on 2 large video databases annotated with emotion labels. 
\\
$\bullet$ We design a video-based face renderer, to decode the parametric representation back to photo-realistic frames. Building upon robust, state-of-the-art face editing techniques (face segmentation, alignment, blending) we modify only the face area, while the background remains unchanged, making it possible to manipulate challenging scenes.
\\
$\bullet$ We conduct extensive qualitative and quantitative experiments, user studies and ablation studies to evaluate our method and compare it with recent state-of-the-art methods. The experiments demonstrate the effectiveness and advantages of our method, which achieves promising results in very challenging scenarios as the ones encountered in movie scenes with moving background objects. 
\\
$\bullet$ We release our code and trained models~\cite{NED_website}.

\section{Related Work}
\label{sec:related}
Face manipulation methods can be divided according to whether they directly edit face portraits through convolutional architectures, or they rely on a geometric face representation:
\\
\textbf{Image-based emotion editing.} The introduction of GANs~\cite{Goodfellow} has sparked a growing line of research in the field of image and video synthesis. The vast majority of works utilise a conditional generator, in the sense that the synthesized image is conditioned on another image (\eg~\cite{pix2pix2017}). This enables translating images between different domains (i.e. \textit{image-to-image translation}) while preserving the content of the source image, even by training on non-parallel data through the idea of \textit{cycle consistency}~\cite{CycleGAN}. The use of such techniques on face images enables the altering of facial attributes (\eg hair color, gender etc.) and constitutes a major part of the so-called area of \textit{deepfakes}. The multi-domain framework of StarGAN~\cite{StarGAN} demonstrated the potential of altering the facial emotions in images by translating them according to the given semantic label (\eg happy, angry etc.). Other techniques make use of continuous emotion labels, such as the \textit{intensity}~\cite{ding2017exprgan}, or the \textit{Valence-Arousal} space~\cite{Lindt2019FacialEE}. Recently, the proposed method of GANmut~\cite{sdapolito2021GANmut} introduced a way of obtaining a 2D interpretable conditional label system even when using a dataset annotated with solely categorical labels of basic emotions.
\\
However, all the above methods translate static frames without taking into account the dynamic nature of facial performance. This is especially essential in the mouth area, as the conveyed speech may be distorted if such techniques are applied independently to every frame of a video. Moreover, they are usually trained on large datasets of images, containing several different identities, which is likely to cause an identity leakage, \eg in cases where a closed mouth is replaced with a smile revealing the teeth of another identity. At the same time, progress in the field shows the potential of generating diverse versions of a given image, by conditioning the generator on dense representations rather than coarse domain labels~\cite{choi2020starganv2}. To overcome the aforementioned limitations, we utilise a GAN-based \textit{domain-to-domain translation} method (inspired by StarGAN v2~\cite{choi2020starganv2}), which translates sequences of subject-agnostic parametric representations of facial expressions instead of images. Then, our person-specific face renderer ensures that the manipulated expressions are synthesized in an identity-preserving way. 
\\
\textbf{Geometry-based face manipulation.} In the last years, the problem of manipulating faces on a parametric space has attracted increased interest. Face reenactment is the most typical example, where the target actor is forced to mimic the expressions of a source subject in a reference video. Some works utilise 2D facial landmarks for capturing the expressions and driving the target actor either via image-warping~\cite{elor2017bringingPortraits} or neural rendering~\cite{zakharov_few-shot_2019}. \textit{3D Morphable Models (3DMMs)}~\cite{Blanz} are a very popular choice, as they offer a disentangled representation of expressions from identity. Traditional techniques~\cite{Thies_real_time},\cite{face2face} perform 3D face reconstruction on the reference video and render the target subject under the source expressions on top of the original target footage. Learning-based methods, like DVP~\cite{DVP} and Head2Head++~\cite{head2head++} use conditional GANs to render the target subject under the given conditions (expressions, pose, eye-gaze).
\\
Nevertheless, these methods offer no semantic control over the generated video, as they directly copy the expressions from a source actor. ICface~\cite{Tripathy_ICface} and FACEGAN~\cite{Tripathy_FACEGAN} present a more intuitive animation framework by conditioning synthesis on \textit{Action Units (AU)} values, but setting individual AU values is a cumbersome process and requires expertise to achieve the desired emotion. Solanki and Roussos \cite{DSM21} train a decoder network that maps \textit{Valence-Arousal} pairs to expression coefficients of a 3D face model, and synthesize the target actor with a neural renderer. Their method, however, totally ignores the original expressions and mouth motion of the actor. Groth \etal~\cite{groth2020altering} try to alter the emotional state of an actor by merely interpolating between his/her expressions and the MoCap data obtained from a reference actor. Kim \etal~\cite{neural_style_preserving_dubbing} presented a style-preserving solution to film dubbing, where the expression parameters of the dubber pass through a \textit{style-translation} network before driving the performance of the foreign actor. Their method preserves the dubber's speech, but can only translate between a pair of speaking-styles (dubber-to-actor). Other methods for generating emotional talking faces include audio-driven~\cite{ji2021audio-driven} and text-driven~\cite{yao2021talkinghead} techniques. To the best of our knowledge, there is no systematic way of translating an existing facial performance in a video to multiple emotions given only semantic information as input, while preserving the original speech. Our method offers an automatic solution to this task through the proposed \textit{3D-based Emotion Manipulator}. It does not attempt to handle the specific speaking styles of two predefined actors (as in \cite{neural_style_preserving_dubbing}), but is able to translate the expressions of any subject to any basic emotion or a given reference style.

\section{Method}
\label{sec:method}

\begin{figure*}[t]
\centering
\includegraphics[trim=0 0 0 0, clip, width=.9\textwidth]{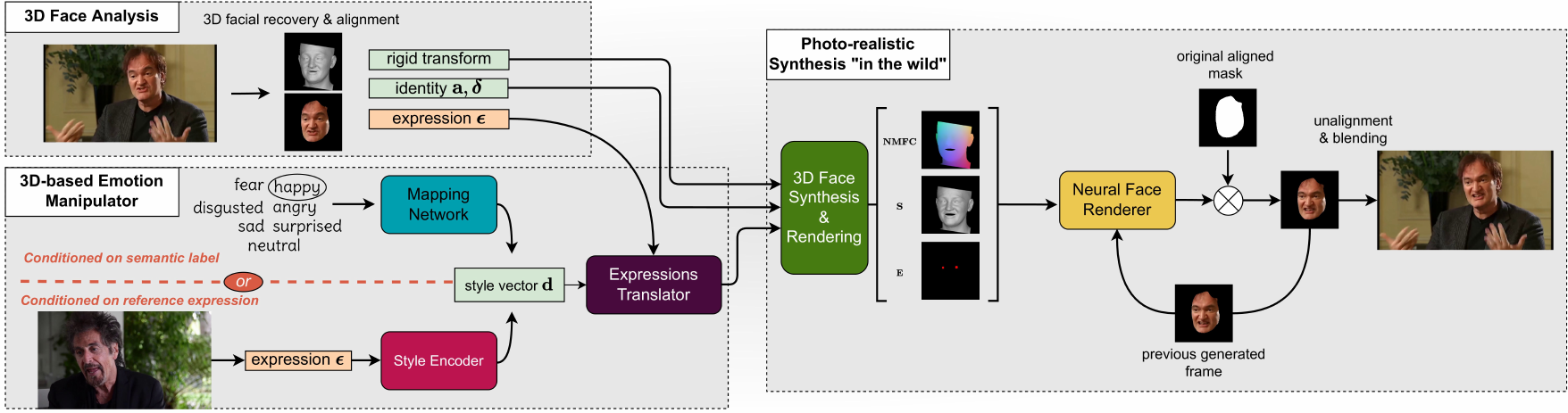}

\caption{Overview of our \textit{Neural Emotion Director (NED)} at inference time. First, we perform 3D facial recovery and alignment on the input frames to obtain the expression parameters of the face. Then, these parameters are translated using our \textit{3D-based Emotion Manipulator}, where the style vector is computed by either a semantic label (i.e., the emotion), or a driving reference video. Finally, the produced 3D facial shape is concatenated with the Normalized Mean Face Coordinate (NMFC) and eye images and fed into a neural renderer (along with previously computed frames), in order to render the manipulated photo-realistic frames.}
\label{fig:arch}
\end{figure*}

Our \textit{Neural Emotion Director (NED)} framework addresses the challenging task of emotion-related semantic manipulation of faces in videos while preserving their speech-related mouth motion. An outline of the proposed pipeline at test time is presented in Fig.~\ref{fig:arch}. It consists of three main modules 
(\textit{3D Face Analysis}, \textit{3D-based Emotion Manipulator} and \textit{Photo-realistic Synthesis ``in the wild''}) that are presented in the following sections. 

\subsection{3D Face Analysis}
\noindent\textbf{Face detection and segmentation:} 
We first perform face detection, cropping and resizing to $256 \times 256$ pixels using \cite{MTCNN}.
We then apply FSGAN \cite{nirkin2019fsgan} to segment the face and remove the background. 
\\
\textbf{3D face reconstruction:} We harness the power of 3DMMs~\cite{Blanz} to estimate the 3D face geometry, while disentangling the expression contributions from the identity-specific and 3D pose ones. This enables us to map the emotion translation problem from the image space to the space of 3D model parameters in a subject-agnostic manner. We perform deep 3D face reconstruction with the recent state-of-the-art method of DECA~\cite{DECA:Siggraph2021} that uses the FLAME model~\cite{FLAME:SiggraphAsia2017}: For each frame of the input video, DECA regresses the parameters of the camera $\textbf{c}\in\mathbb{R}^{3}$, head pose $\textbf{p}\in\mathbb{R}^{6}$ (including 3 jaw articulation parameters), identity $\textbf{a}\in\mathbb{R}^{100}$, expression $\textbf{e}\in\mathbb{R}^{50}$, as well as the person-specific detail vector  $\boldsymbol{\delta}\in\mathbb{R}^{128}$, which adds mid-frequency details to the face geometry. We use the latter to create detailed shape images $\textbf{S}\in\mathbb{R}^{256\times256\times3}$ (see 3D facial shapes in Fig.~\ref{fig:arch}).



\noindent\textbf{Landmark detection and face alignment:} We use FAN~\cite{bulat2017far} to obtain 68 facial landmarks for each frame. Afterwards, similarly to \cite{head2head++}, we estimate eye pupil coordinates based on the inverse intensities of the pixels within the eye area and create eye images $\textbf{E}\in\mathbb{R}^{256\times256\times3}$ that provide the face renderer with information about the eye-gaze. However, in contrast to \cite{head2head++}, we only draw two red disks around eye pupils and not the edges of the outline. This is because we accurately integrate information about eye blinking within the NMFC and detailed shape images (see Sec.~\ref{sec:renderer}), thanks to the reliable reconstructions in the eye regions obtained by DECA~\cite{DECA:Siggraph2021}. 
We then align all face frames to a face template, based on the extracted face landmarks and Procrustes analysis. We found that such face alignment boosts our face renderer's generalization ability. We apply the same alignment to the NMFC, shape, and eye images. For more details on the face alignment step, please refer to the Supp.~Material. 

\subsection{3D-based Emotion Manipulator}
Following the \textit{3D Face Analysis} step, information related to the facial expression in a frame is encoded in the expression vector $\textbf{e}\in\mathbb{R}^{50}$ and the 3 jaw parameters (formed as the last 3 components of the pose vector $\textbf{p}\in\mathbb{R}^{6}$). 
We expand the 50 expression parameters $\textbf{e}$ with the jaw opening $\textbf{p}_4$ (the $1^{\text{st}}$ jaw articulation parameter), as this is the main parameter that describes speech-related mouth motions. Thus, we concatenate the two into a single vector, namely the full expression vector $\boldsymbol{\epsilon}=(\textbf{p}_4;\textbf{e})\in\mathbb{R}^{51}$, 
henceforth called \textit{expression vector} for simplicity. To cope with the dynamic nature of facial expressions, we group frames into $N$-length sequences $\textbf{s}$=$(\boldsymbol{\epsilon}^n,..,\boldsymbol{\epsilon}^{n+N-1})$, with $N$=10. Given the set ${\cal Y}$ of the $c$=7 emotion labels (neutral, happy, fear, sad, surprised, angry, disgusted), each denoting a distinct domain, and the set ${\cal S}$ of sequences of expression parameters, we design a \textit{3D-based Emotion Manipulator} that  translates a sequence of expression vectors $\textbf{s}$$\in$$\cal S$ to a given emotion $y$$\in$$\cal Y$ in a realistic way that preserves the original mouth motion. 
Inspired by the StarGAN v2~\cite{choi2020starganv2} framework, which offers diversity in the generated samples by conditioning the generator on a continuous style vector, 
we design an architecture with the following four modules:
\\
\textbf{Expressions Translator:} The translator $G$ takes as input a sequence of expressions $\textbf{s}$ and a style vector $\textbf{d}\in\mathbb{R}^{16}$ and translates $\textbf{s}$ into an output sequence of expression vectors $G(\textbf{s}, \textbf{d}) \in \cal S$ that reflects the speaking style encoded in $\textbf{d}$. To inject $\textbf{d}$ into $G$, we concatenate $\textbf{d}$ with each of the $N$ vectors of the sequence.
\\
\textbf{Style encoder:} Our style encoder $E$ extracts the emotion-related style vector $\textbf{d}=E(\textbf{s})$ of an input sequence $\textbf{s}$ and, thus, enables the translator $G$ to translate a given sequence according to the speaking style extracted from a reference sequence. In contrast to~\cite{choi2020starganv2}, our style encoder does not require any knowledge about the ground truth emotion label $y$ of the reference sequence $\textbf{s}$.
\\
\textbf{Mapping network:} The mapping network $M$ learns to generate style vectors $\textbf{d}=M_y(\textbf{z}) \in \mathbb{R}^{16}$ related to a target emotion $y \in \cal Y$, by transforming a latent code $\textbf{z} \in \mathbb{R}^{4}$ sampled from a normal distribution. Here, $M_y(\cdot)$ denotes the output branch of $M$ that corresponds to the emotion $y$. This network allows the translator to translate a sequence of expressions to a target emotion, by merely sampling random noise, and specifying the desired semantic emotion label.
\\
\textbf{Expressions Discriminator:} Our discriminator $D$ has $c=7$ branches (similarly to $M$) and learns to discriminate between real $\textbf{s}$ and fake $G(\textbf{s}, \textbf{d})$ sequences of each domain $y$ by outputting a scalar value $D_y(\textbf{s})$ for each branch.
\\
The network $M$ follows a simple MLP architecture, whereas $G,E$ and $D$ use recurrent architectures with LSTM units~\cite{LSTM}.

\subsubsection{Training and testing of the Emotion Manipulator} \label{sec:TrainEmManip}
\begin{figure}[h]
\centering
\includegraphics[width=0.45\textwidth]{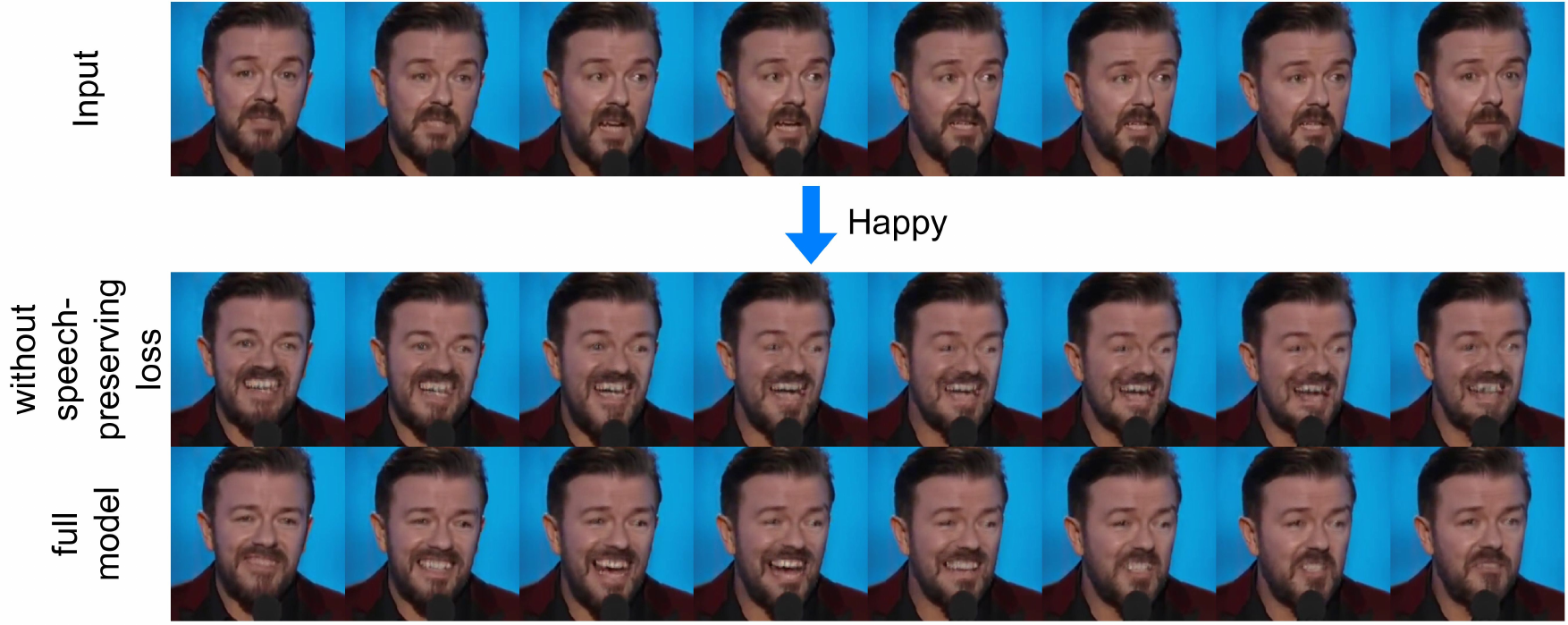}
\caption{Effect of our speech-preserving loss. Without this loss (middle row), the result does not preserve the mouth movement from the input video. In contrast, our full model (bottom row) translates the expression of the actor to happy while preserving his mouth movements and speech. Please zoom in for details.}
\label{fig:mouth_loss}
\end{figure}

Given a dataset of sequences of expression vectors $\textbf{s} \in \cal S$ and their corresponding ground truth labels of emotions $y \in \cal Y$, we train our networks in 2 alternating steps: 1) first we sample $\textbf{z} \in \mathbb{R}^{4}$ from a normal distribution, we randomly pick a target domain $\tilde{y} \in \cal Y$, and employ our mapping network for generating the speaking style $\tilde{\textbf{d}}=M_{\tilde{y}}(\textbf{z})$. 2) then we directly extract the style from a reference sequence $\tilde{\textbf{s}}$ with our style encoder $\tilde{\textbf{d}}=E(\tilde{\textbf{s}})$ and store the reference label $\tilde{y}$. In both cases, the translator combines an input sequence $\textbf{s}$ (belonging to domain $y$) with the style vector $\tilde{\textbf{d}}$ and produces an output sequence $G(\textbf{s}, \tilde{\textbf{d}})$ belonging to the target domain $\tilde{y}$ and resembling the speaking style in $\tilde{\textbf{d}}$. The networks are then updated using the following objectives.
\\
\textbf{Adversarial loss:} We use LSGAN~\cite{LSGAN} with labels $b$=$c$=1 for real samples and label $a$=0 for fake ones. This way the mapping network $M$ learns to output the speaking styles that belong to the emotional domain $\tilde{y}$ and the translator to produce sequences of the target domain that are indistinguishable from the real ones.
\\
\textbf{Style reconstruction loss:} As in~\cite{choi2020starganv2}, we make sure the output sequence reflects the given style by using a loss that enforces the style vector of the translated sequence, as extracted by the style encoder $E$, to  match the desired one.
\\
\textbf{Cycle consistency loss:} We use the cycle consistency loss~\cite{CycleGAN,StarGAN}, which encourages the translator to produce sequences that preserve the content of the input sequence, so that the input sequence can be reconstructed by translating the output sequence back to the original style $\hat{\textbf{d}}=E(\textbf{s})$, as extracted by $E$.
\\
\textbf{Speech-preserving loss:} As observed in~\cite{neural_style_preserving_dubbing}, the cycle consistency loss does not always guarantee that the original mouth motion related to speech is preserved by the translator. To this end, we leverage our carefully selected FLAME model~\cite{FLAME:SiggraphAsia2017}, which explicitly controls the mouth opening through the $1^{\text{st}}$ jaw parameter. Thus, we add an extra constraint to the total objective, that takes into account only this mouth related parameter, instead of the whole expression vector as in~\cite{neural_style_preserving_dubbing}. 
By properly defining this objective as the maximization of the \textbf{correlation} between the original and the translated jaw opening variable, we manage to balance our challenging and contradictive goal of altering the emotion without distorting the perceived speech (see Fig.~\ref{fig:mouth_loss}).
\\
\textbf{Overall objectives:} The objective for $G, E$ and $M$ corresponds to a weighted summation of the Adversarial, Style reconstruction, Cycle consistency and Speech-preserving losses. 
The objective for $D$ corresponds to the discriminator loss. 
More details and mathematical formulas for the adopted loss functions can be found at the Supp.~Material. 

We train our \textit{3D-based Emotion Manipulator} on two video databases with annotations of the 6 basic emotions plus neutral: the Aff-Wild2 database~\cite{kollias2019deep, kollias2019expression, kollias2019face, kollias2020analysing, kollias2021affect, kollias2021analysing, kollias2021distribution, zafeiriou2017aff} of ``in-the-wild" videos and the MEAD database~\cite{wang2020mead} (we exclude \textit{contempt} for MEAD to match the emotions in Aff-Wild2). 
We recover the expression parameters for every frame of the videos and extract sliding windows of length $N$. 
To get the best of the two databases, we pre-train our networks in Aff-Wild2 and then fine-tune them on a subset of MEAD.

During \textbf{testing}, to transform the expressions of a whole input video, we slide the $N$-length window by 1 frame at a time, translate the sequence through $G$, and use a weighted averaging of Gaussian type to handle overlaps. The conditional style vector is either generated by $M$ by choosing a target emotion or extracted from a reference video of arbitrary length by $E$. In the latter case, we process the whole reference video sequentially, extracting 
a sequence of style vectors adopting the same sliding pattern. We then take the \textbf{geometric median}~\cite{L1_median} of them as the style vector representing the whole reference video and feed it to $G$.

\subsection{Photo-realistic Synthesis ``in the wild''}
\label{sec:renderer}
\noindent\textbf{3D Face Synthesis \& Rendering:} Having modified the expression parameters through our \textit{3D-based Emotion Manipulator}, we synthesize a manipulated 3D face geometry under the new emotion. We then render it (using conventional 3D graphics) to a convenient representation for neural rendering, the so-called 
\textit{Normalized Mean Face Coordinate (NMFC)} image \cite{head2head++}, and concatenate it with the similarly-rendered detailed shape image $\textbf{S}$ and eye image $\textbf{E}$. 
\\
\textbf{Neural Face Renderer:} We feed our neural renderer with the  \textbf{NMFC}, $\textbf{S}$ and $\textbf{E}$ images as conditional input. We build it upon the publicly available Head2Head++~\cite{head2head++} implementation and train it on the training footage of a target actor in a self-reenactment fashion (i.e. with the original face geometry). 
We follow the recurrent scheme of~\cite{head2head++} by feeding the generator with the conditional inputs of both the current as well as the two previous frames, along with the two previously generated images. However, in contrast to \cite{head2head++}, 
we include $\textbf{S}$ as additional conditional input, as described above, and constrain image synthesis to the aligned and masked faces, 
since we account for changing background. As in~\cite{head2head++} we employ a dedicated \textit{mouth discriminator} to enhance realism in the mouth area. 
\\
\textbf{Blending:} 
Face alignment is reversed by transforming the generated images according to the inverse of the previously stored alignment matrix. 
We then carefully blend the synthesized face with the original background, enabling the manipulation of ``in-the-wild'' videos. 
For this, we use multi-band blending~\cite{burt_image_mosaics}, as we found it to perform better than soft-masking or Poisson editing~\cite{poisson_editing} in terms of smooth boundary transition. 
\\
For more details on this module, please refer to the Supp.~Material.

\section{Experimental Results}
We conduct comprehensive qualitative and quantitative evaluations of our method and comparisons with recent state-of-the-art methods. \textbf{Additional results and visualizations are provided on our website}~\cite{NED_website}.

Our experiments use the following \textbf{datasets}: \textbf{YouTube Actors dataset:} We collected a small dataset from 6 YouTube videos that included facial videos of 6 actors during film scenes, TV shows and interviews. 
\textbf{MEAD dataset:} We chose 3 actors from the recent MEAD database~\cite{wang2020mead}. 
For every actor, we selected 30 videos for each 
of the 6 basic emotions (happy, angry, surprised, fear, sad, disgusted) plus neutral, resulting to a total of 630 videos from MEAD. For more details about the datasets used in the experiments, please refer to the Supp.~Material. 

A person-specific face renderer had to be trained separately for each actor of these datasets. Yet, we found that by training a new model from scratch for a given actor, the generator tends to overfit to this actor's idiosyncrasies failing to synthesize novel emotions, if those were not present in his/her training footage. 
For YouTube actors, finding a few-minute-long footage that covers the total range of emotions is often impossible. To overcome the challenge of generating unseen expressions of an actor while preserving his/her identity we propose training a single \textbf{meta-renderer} on a mix of videos including YouTube and MEAD actors, and then, fine-tuning the \textbf{meta-renderer} independently for each actor. This helps by sharing the expressivity of the MEAD actors among the YouTube actors. For a further explanation of this process please refer to the Supp.~Material.

We \textbf{compare} our method with the following recent methods:\textbf{ GANmut~\cite{sdapolito2021GANmut}}, which transforms an input cropped facial image according to a 2D continuous emotion label. The transformed image is then placed in its original position in the full frame. For a fair comparison, we use the 2D vectors that correspond to one of the pure 7 emotions with maximum intensity. Also, we apply the method in every frame of the input video. 
\textbf{ICface~\cite{Tripathy_ICface}}, which also transforms an input cropped image, but based on the determination of pose and AU values. For a fair comparison, this method is only included in the self-reenactment experiments, where the ground truth is available and AU values can be extracted from it in a well-defined and consistent manner. 
\textbf{DSM \cite{DSM21}}, which performs semantic manipulation of facial videos based on a categorical emotion label. The labels that this method supports are neutral and only 4 basic emotions (happy, surprised, fear, sad), which is a subset of the labels considered by the other methods and thus the relevant comparisons with DSM are done in this subset. 
It is worth mentioning that the literature of photo-realistic emotion manipulation of faces in videos is still extremely limited and the aforementioned 3 methods were the only ones for which we could find source code to run for our evaluations. Details about running our method and the methods we compare with can be found at the Supp.~Material. 

\begin{figure}[h]
\centering
\includegraphics[trim=0 0 0 0, clip, width=0.45\textwidth]{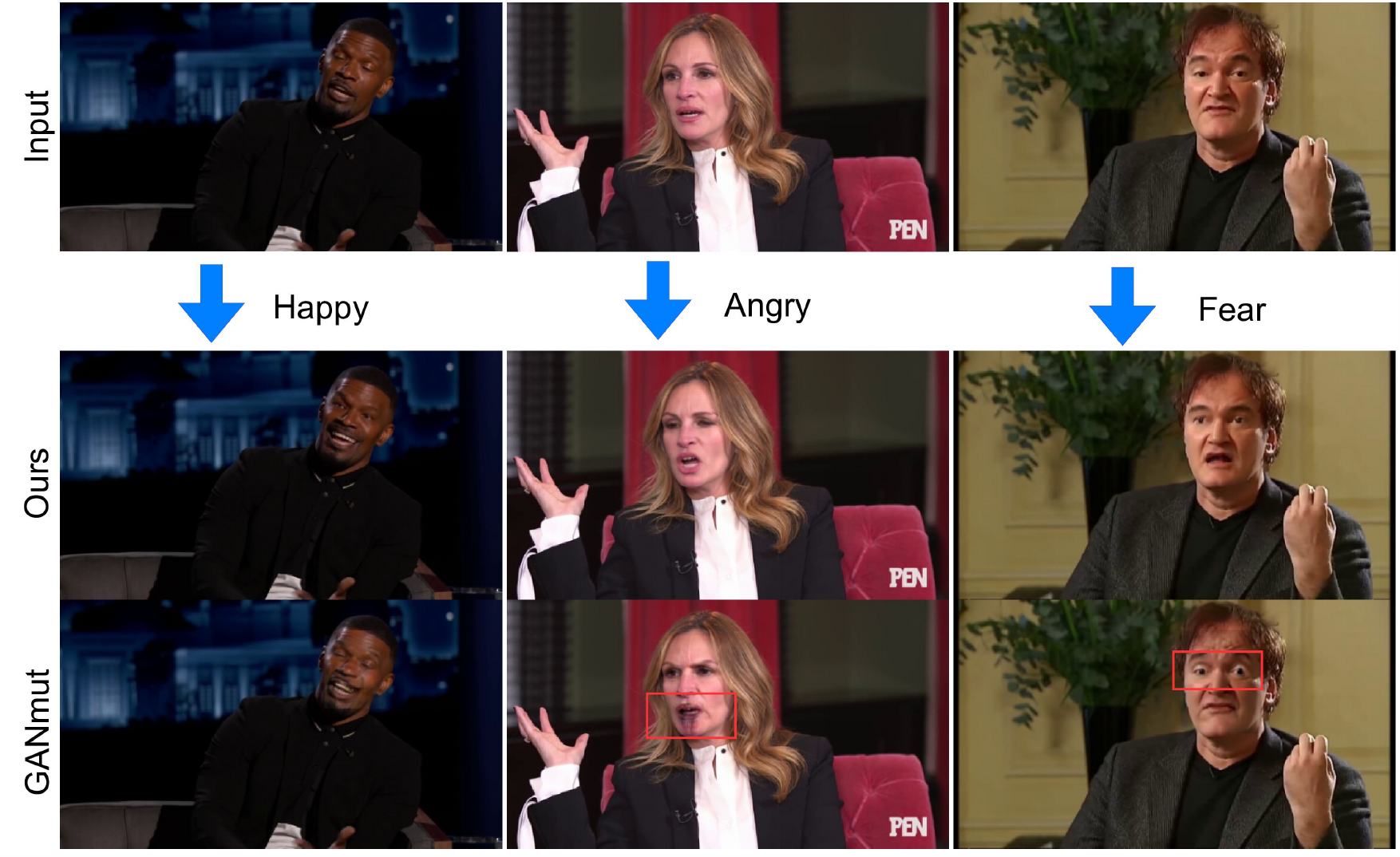}
\caption{Qualitative comparison between our method and GANmut~\cite{sdapolito2021GANmut}. We show examples of 3 emotions for 3 actors of the YouTube Actors dataset. GANmut produces visible artifacts especially in the mouth and eyes areas. Please zoom in for details.}
\label{fig:qualitative}
\vspace{-.4cm}
\end{figure}

\subsection{Quantitative Comparisons}

\begin{figure}[t]
\centering
\includegraphics[width=0.45\textwidth]{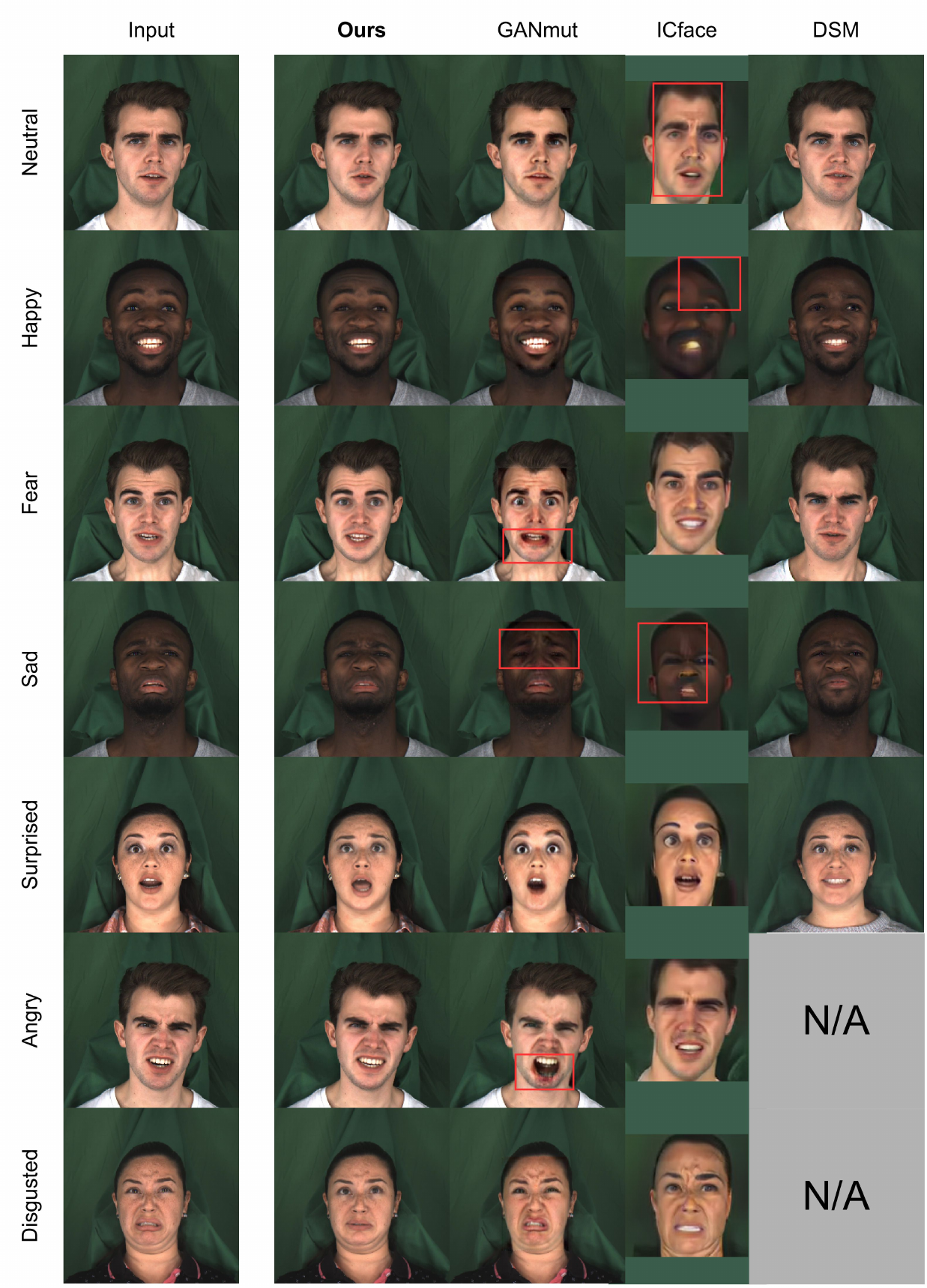}
\caption{Visual comparison with state-of-the-art methods in the emotional ``self-translation" experiment on the MEAD actors. Note that ICface~\cite{Tripathy_ICface} requires a tighter face cropping and padding with the background color has been used for visualization.}
\label{fig:quantitative_comp}
\vspace{-0.4cm}
\end{figure}

\begin{table}
\setlength{\tabcolsep}{4pt}
\footnotesize
\label{tabl:abl1}
\centering
\begin{tabular}{l|cc|cc|cc|cc}
{} & \multicolumn{2}{c|}{Ours} & \multicolumn{2}{c|}{GANmut\cite{sdapolito2021GANmut}} & \multicolumn{2}{c|}{DSM\cite{DSM21}} & \multicolumn{2}{c}{ICface\cite{Tripathy_ICface}} \\
{} & FAPD & FID &   FAPD & FID &   FAPD &  FID &  FAPD &  FID \\
\hline
neutral   & \textbf{14.9} & \textbf{2.1} &   16.8 & 2.9 &   22.1 &  5.5 &   40.0 & 45.8 \\
happy     & 17.8 & 3.4 &   \textbf{15.0} & \textbf{2.6} &   27.0 & 10.3 &   43.7 & 50.5 \\
fear      & \textbf{18.4} & \textbf{3.0} &   20.5 & 4.3 &   28.0 &  8.5 &   43.0 & 46.6 \\
sad       & 19.0 & \textbf{3.0} &   \textbf{18.1} & 4.3 &   24.5 &  8.8 &   38.6 & 47.4 \\
surprised & \textbf{18.9} & \textbf{2.9} &   19.1 & 7.2 &   27.3 & 11.8 &   46.1 & 45.0 \\
angry     & \textbf{18.4} & \textbf{3.0} &   22.4 & 4.2 &    - &  - &   51.7 & 53.5 \\
disgusted & 18.1 & \textbf{3.1} &   \textbf{15.1} & 4.5 &    - &  - &   39.7 & 54.6 \\ 
\hline
avg. (7) & \textbf{17.9} & \textbf{2.9} &   18.1 & 4.3 &   - &  - &   43.3 & 49.0 \\
avg. (5) & \textbf{17.8} & \textbf{2.9} &   17.9 & 4.3 &   25.8 &  9.0 &   42.3 & 47.1 \\
\end{tabular}  
\caption{Quantitative comparisons on MEAD in the emotional ``self-translation" experiment. Bold values denote the best value for each metric (lower is better). Averaging is done over both the full set of 7 emotion labels and the set of 5 labels supported by DSM~\cite{DSM21}, for the sake of fair comparison.}
\label{tab:quant_FID}
\vspace{-0.5cm}
\end{table}

To assess the performance of each method in manipulating the emotion we use a variant of the self-reenactment task that includes the manipulation of the expressions. In particular, given a video of emotional label $y$ (\eg happy), we modify its emotion by defining the target emotion as the same label $y$ (from happy to happy). In this case, the output video should match the original one. Specifically, the discrepancy between the ``self-translated" and the real video is measured using the following metrics: 1) \textbf{Face Average Pixel Distance (FAPD):} which is the mean
$\ell_2$ distance of RGB values across all facial pixels and frames, between the ground truth and generated video. We use the extracted face mask to define the face area. 2) \textbf{Frech\'{e}t inception distance (FID)\cite{heusel2017fid}:} which is computed by using the feature vectors from a state-of-the-art face recognition network\cite{deng2018arcface} for all the ground truth and the generated frames.

Our \textit{3D-based Emotion Manipulator} can translate the sequences to the same domain by simply using their own style vector as extracted by the style encoder $E$. However, for the other 3 methods, the label $y$ of the original video has to be known so that it can be used as the target label. Therefore, our quantitative comparison is performed in our \textbf{MEAD dataset}, in which videos are emotionally annotated by the authors of~\cite{wang2020mead}. Specifically, for each actor we use 4 videos per emotion, leading to a total of 84 videos, of average duration $\sim3$ secs. Results are presented in Tab.~\ref{tab:quant_FID}. For a visual comparison please refer to Fig.~\ref{fig:quantitative_comp}. Note that \textit{disgusted} and \textit{angry} are not supported by the DSM method.

As can be seen, our method outperforms the baselines in both metrics overall. We also exhibit superior performance in almost all 7 emotions individually. This shows the higher realism of our synthesized videos as well as the better expression transferability in terms of identity preservation (see Fig.~\ref{fig:quantitative_comp} for artifacts produced by GANmut and ICface).

\subsection{User Studies}
\noindent We also conducted two web-based user studies:
\\
\textbf{Emotion Recognition and Realism on MEAD Database:} 
In the first user study, participants were shown randomly shuffled manipulated videos of 3 actors from MEAD database in all 6 basic emotions and were asked to rate the realism of the footage on a Likert 5-point scale, as well as recognize the emotion shown (from a drop-down list including all 6 emotions). Apart from our method, the questionnaire showed to the participants manipulated videos by GANmut~\cite{sdapolito2021GANmut}, DSM~\cite{DSM21}, as well as the original real videos from MEAD. 
In total the questionnaire included 66 videos and 20 participants completed it. 
The results can be seen in Tab.~\ref{tab:user_study_MEAD} where we observe that all methods have relatively low realism scores. This can be attributed to the fact that the real videos in MEAD include particularly intense expressions, which probably resulted to an overall low frequency of ratings of 4 or 5 (even for real videos) and to an increased tendency to use these ratings  (whenever they were used) more exclusively for real videos. 
However, we see that our method achieves significantly higher realism scores than the other methods, consistently across all 6 emotions. 
In terms of emotion recognition accuracy, we observe that our method synthesized videos with manipulated emotions that were consistently easier to be recognised by the participants, in comparison to DSM. However, this is not the case when we compare our method with GANmut: The synthetic videos of GANmut achieved a very high accuracy rate, which is even higher than the accuracy for real videos. This, in combination with the low realism score of GANmut (see also the ``exaggerated'' emotions in Figs.~\ref{fig:qualitative} and \ref{fig:quantitative_comp}), reflects the fact that GANmut synthesizes intense expressions that typically look fake but are easily recognizable.
\\
\textbf{Realism on YouTube Actors:} In the second study, we presented users with manipulated videos (including the original audio) of 6 YouTube actors in all 6 basic emotions and asked them to rate the realism of the footage, following the same protocol as in the first study. We did not evaluate emotion recognition in this study since ground truth emotion annotations do not exist for this dataset (to compare with).
The study included a random shuffling of videos manipulated by our method and GANmut, as well as the original videos. For some indicative frames of these videos, please refer to Fig.~\ref{fig:qualitative}. DSM was not used for this study, since it cannot handle dynamic backgrounds such as those found in YouTube videos. The questionnaire included 54 videos in total and was completed by 50 participants. The ratings obtained can be seen in Tab.~\ref{tab:user_study_youtube}. 
We observe that the realism scores for both methods are relatively low, which can be attributed to the highly challenging task of manipulating the emotions in videos, especially under ``in-the-wild'' conditions as is the case for the YouTube Actors dataset. 
However, our method achieves a better score than GANmut 
and for example succeeds in synthesizing realistic videos more than 20\% of the times for 3 out of 6 actors, which is a promising result that shows the potential of our approach.
Furthermore, in terms of the most frequent rating, we see that our method is consistently better than GANmut, as it yields a rating of 3 or 2 as the most frequent answer for almost all actors, in contrast to GANmut that yields the rating of 1 as the most frequent.

\begin{table}[h!]
\scriptsize
\centering
\setlength{\tabcolsep}{1.8pt}

\begin{tabular}{l|cccc|cccc}
{} & \multicolumn{4}{c|}{Realism} & \multicolumn{4}{c}{Accuracy} \\
{} & Ours & GANmut & DSM & Real Videos & Ours & GANmut & DSM & Real Videos \\
\hline
happy     &   17\% &      3\% &      8\% &           80\% &   63\% &     90\% &     42\% &           90\% \\
fear      &   32\% &      7\% &     10\% &           67\% &   33\% &     75\% &     13\% &           25\% \\
sad       &   30\% &     18\% &     12\% &           55\% &   13\% &     78\% &     25\% &           65\% \\
surprised &   22\% &      8\% &      7\% &           82\% &   17\% &     82\% &      5\% &           82\% \\ 
angry     &   25\% &     10\% &      - &           78\% &   50\% &     98\% &     - &           80\% \\
disgusted &   40\% &     20\% &      - &           67\% &   33\% &     40\% &      - &           60\% \\\hline
avg.   &   28\% &     11\% &      9\% &           71\% &   35\% &     77\% &     21\% &           67\% \\
\end{tabular}
   \caption{Realism ratings (percentage  of  users  that  rated  the videos  with  4  or  5) and classification accuracy of the user study on MEAD (c.f.~Supp.~Material for detailed scores).}
   \label{tab:user_study_MEAD}
   \vspace{-0.3cm}
\end{table}

\begin{table}[h!]
\setlength{\tabcolsep}{1.5pt}
\scriptsize
\centering
\begin{tabular}{l|cccccc|cccccc|cccccc}
{} & \multicolumn{6}{c|}{Ours} & \multicolumn{6}{c|}{GANmut\cite{sdapolito2021GANmut}} & \multicolumn{6}{c}{Real Videos} \\
{} &    1 &  2 &  3 &  4 &  5 & `real' &      1 &  2 &  3 &  4 &   5 & `real' &            1 &   2 &  3 &  4 &   5 & `real' \\
\hline
McDormand   &   32 & 32 & \textbf{52} & 21 & 13 &           23\% &     \textbf{59} & 23 & 16 & 28 &  24 &           35\% &          0 &   2 &  3 & 21 & \textbf{124} &           97\% \\
Pacino    &   19 & 45 & \textbf{53} & 25 &  8 &           22\% &     40 & \textbf{41} & 26 & 24 &  19 &           29\% &          0 &   3 &  6 & 29 & \textbf{112} &           94\% \\
Tarantino &   \textbf{70} & 29 & 23 & 17 & 11 &           19\% &     \textbf{72} & 20 & 26 & 19 &  13 &           21\% &            1 &   5 &  9 & 43 &  \textbf{92} &           90\% \\
\!\!\!\!\!McConaughey   &   37 & \textbf{63} & 33 & 13 &  4 &           11\% &     \textbf{88} & 43 & 12 &  7 & 0 &            5\% &          0 &   4 & 18 & 33 &  \textbf{93} &           85\% \\
Roberts     &   34 & \textbf{60} & 39 & 12 &  5 &           11\% &     \textbf{88} & 27 & 17 & 13 &   5 &           12\% &          0 & 0 &  3 & 24 & \textbf{123} &           98\% \\
Foxx     &   26 & 35 & \textbf{39} & 34 & 15 &           33\% &     \textbf{79} & 43 & 18 &  6 &   4 &            7\% &          0 & 0 &  7 & 31 & \textbf{111} &           95\% \\ \hline
avg.   &   36 & \textbf{44} & 40 & 20 &  9 &           20\% &     \textbf{71} & 33 & 19 & 16 &  13 &           18\% &            1 &   4 &  8 & 30 & \textbf{109} &           93\% \\
\end{tabular}
 \caption{Realism ratings of the user study on 6 YouTube actors. Columns 1-5 show the number of times that users gave this rating. The column ``real" shows the percentage of users that rated the videos with 4 or 5. Bold values denote the most frequent user rating for each method and actor.}
 \label{tab:user_study_youtube}
 \vspace{-0.5cm}
\end{table}

\subsection{Ablation Study}
In order to experimentally confirm our design choices for our face renderer,
we performed an ablation study in a pure self-reenactment setting, without any manipulation of the expressions with our \textit{3D-based Emotion Manipulator}. Specifically, we randomly chose 3 actors from our \textbf{YouTube dataset} and trained 4 different variations of our renderer from scratch (i.e. without the meta-renderer): First, we \textbf{omitted the detailed shape images} $\textbf{S}$ as extra conditioning input. 
Second, we \textbf{omitted the face alignment step}.
Third, we trained our \textbf{full model}, but \textbf{without the meta-renderer} stage as in all previous variations. Finally, we also considered our \textbf{full model including the meta-renderer}. 

The performance of each variation in relation to the ground truth frames was calculated by means of the FAPD and FID (as defined above), as well as 2 more metrics, namely the APD (same as FAPD but calculated on the entire image) and Mouth-APD (MAPD, same as FAPD but calculated only on a 72$\times$72 pixel area around the mouth center). The results for all metrics, as shown in Tab.~\ref{tab:ablation}, demonstrate the contribution of both the detailed shape images and the face alignment, especially in the highly challenging mouth area. Finally, the metrics of the fourth row reveal that the meta-renderer improves the results even further. 

\begin{table}[h]
\scriptsize
\centering
 \begin{tabular}{c|cccc}
    Variations & APD & FAPD & MAPD & FID\\
    \hline
    w/o detailed shape images & 5.01 & 12.57 & 13.27 & 5.06\\
    w/o face alignment & 5.01 & 12.64 & 14.59 & 4.77\\
    full model & \underline{4.63}  & \underline{11.35} & \underline{12.20} & \underline{4.53}\\
    \hline
   full model \textbf{with meta-renderer}& {\textbf{4.49}} & {\textbf{10.89}} & {\textbf{11.66}} & {\textbf{4.38}}\\
  \end{tabular}
 \caption{Ablation study results under the self reenactment setting, averaged across three YouTube actors from our dataset. For all metrics, lower values indicate better performance. Bold and underlined values correspond to the best and the second-best value of each metric, respectively.}
\label{tab:ablation}
\vspace{-0.5cm}
\end{table}

\section{Discussion}

We have shown promising results in various scenarios for a novel application of neural rendering. As expected, there are still limitations in our approach, which can pave the way for future work. For example, our renderer produces images of medium resolution which may lead to blurry faces when blending them with high-resolution backgrounds. This may explain some of the low realism scores reported in our user studies. Extending the method to a higher resolution could facilitate the integration of such techniques to the film industry. Successful approaches have already been shown for \textbf{face swapping} via progressive training \cite{HighResolutionNeuralFaceSwapping}. Also, to further improve the realism of emotion manipulation, the audio content should be modified in a similar way. However, state-of-the-art results in \textit{emotional voice conversion}~\cite{Zhou2021EmotionalVC} are still far from matching the quality of synthetic visual content. 
\\
\textbf{Note on social impact}.~Despite their positive impact, deep learning systems for video manipulation have raised concerns related to the distribution of fake news and other negative social impact \cite{chesney2019deep,diakopoulos2021anticipating,yadlin2021whose,johnson2021deepfakes}. While our goal of speech preservation is inherently opposite to most deepfakes where the output combination of a person and its utterances is entirely fake, our method could also be misused in cases where the conveyed meaning heavily depends on the apparent emotion (\eg political speeches). We believe that scientists working in the relevant fields need to seriously take into account these risks and ethical issues. Some of the countermeasures include contributing in raising public awareness about the capabilities of current technology and developing systems that detect deepfake videos \cite{tolosana2020deepfakes,mirsky2021creation,yu2021survey}.

\section{Conclusion}
We proposed \textit{Neural Emotion Director (NED)}, a novel approach for photo-realistic manipulation of the emotions of actors in videos. Our new \textit{3D-based Emotion Manipulator} translates facial expressions by carefully preserving the speech-related content of the source performance, while our \textit{Photo-realistic Synthesis} module faithfully synthesizes the target actor's face and composites it onto the original video. Our extensive experimental results demonstrate the advantages of our framework over recent state-of-the-art methods and its effectiveness under ``in-the-wild'' conditions.
\\
\textbf{Acknowledgments.} A. Roussos was supported by the Hellenic Foundation for Research and Innovation (HFRI) under the ``$1^{st}$ Call for HFRI Research Projects to support Faculty members and Researchers and the procurement of high-cost research equipment'' Project I.C.Humans, Number: 91.

{\small
\bibliographystyle{ieee_fullname}
\bibliography{egbib}

\begin{thebibliography}{10}\itemsep=-1pt

\bibitem{NED_website}
\url{https://foivospar.github.io/NED/}.

\bibitem{elor2017bringingPortraits}
Hadar Averbuch-Elor, Daniel Cohen-Or, Johannes Kopf, and Michael~F. Cohen.
\newblock Bringing portraits to life.
\newblock {\em ACM Transactions on Graphics (Proceeding of SIGGRAPH Asia
  2017)}, 36(6):196, 2017.

\bibitem{Blanz}
Volker Blanz and Thomas Vetter.
\newblock A morphable model for the synthesis of 3d faces.
\newblock In {\em Proceedings of the 26th Annual Conference on Computer
  Graphics and Interactive Techniques}, SIGGRAPH '99, page 187–194, USA,
  1999. ACM Press/Addison-Wesley Publishing Co.

\bibitem{bulat2017far}
Adrian Bulat and Georgios Tzimiropoulos.
\newblock How far are we from solving the 2d \& 3d face alignment problem? (and
  a dataset of 230,000 3d facial landmarks).
\newblock In {\em Proceedings of the IEEE International Conference on Computer
  Vision (ICCV)}, 2017.

\bibitem{burt_image_mosaics}
Peter~J. Burt and Edward~H. Adelson.
\newblock A multiresolution spline with application to image mosaics.
\newblock {\em ACM Trans. Graph.}, 2(4):217–236, Oct. 1983.

\bibitem{chesney2019deep}
Bobby Chesney and Danielle Citron.
\newblock Deep fakes: A looming challenge for privacy, democracy, and national
  security.
\newblock {\em Calif. L. Rev.}, 107:1753, 2019.

\bibitem{StarGAN}
Yunjey Choi, Minje Choi, Munyoung Kim, Jung-Woo Ha, Sunghun Kim, and Jaegul
  Choo.
\newblock Stargan: Unified generative adversarial networks for multi-domain
  image-to-image translation.
\newblock In {\em Proceedings of the IEEE Conference on Computer Vision and
  Pattern Recognition (CVPR)}, 2018.

\bibitem{choi2020starganv2}
Yunjey Choi, Youngjung Uh, Jaejun Yoo, and Jung-Woo Ha.
\newblock Stargan v2: Diverse image synthesis for multiple domains.
\newblock In {\em Proceedings of the IEEE Conference on Computer Vision and
  Pattern Recognition (CVPR)}, 2020.

\bibitem{sdapolito2021GANmut}
Stefano d'Apolito, Danda~Pani Paudel, Zhiwu Huang, Andres Romero, and Luc
  Van~Gool.
\newblock Ganmut: Learning interpretable conditional space for gamut of
  emotions.
\newblock In {\em Proceedings of the IEEE Conference on Computer Vision and
  Pattern Recognition (CVPR)}, 2021.

\bibitem{deng2018arcface}
Jiankang Deng, Jia Guo, Xue Niannan, and Stefanos Zafeiriou.
\newblock Arcface: Additive angular margin loss for deep face recognition.
\newblock In {\em Proceedings of the IEEE International Conference on Computer
  Vision and Pattern Recognition (CVPR)}, 2019.

\bibitem{diakopoulos2021anticipating}
Nicholas Diakopoulos and Deborah Johnson.
\newblock Anticipating and addressing the ethical implications of deepfakes in
  the context of elections.
\newblock {\em New Media \& Society}, 23(7):2072--2098, 2021.

\bibitem{ding2017exprgan}
Hui Ding, Kumar Sricharan, and Rama Chellappa.
\newblock Exprgan: Facial expression editing with controllable expression
  intensity.
\newblock {\em AAAI}, 2018.

\bibitem{head2head++}
Michail~Christos {Doukas}, Mohammad~Rami {Koujan}, Viktoriia {Sharmanska},
  Anastasios {Roussos}, and Stefanos {Zafeiriou}.
\newblock Head2head++: Deep facial attributes re-targeting.
\newblock {\em IEEE Transactions on Biometrics, Behavior, and Identity
  Science}, 3(1):31--43, 2021.

\bibitem{DECA:Siggraph2021}
Yao Feng, Haiwen Feng, Michael~J. Black, and Timo Bolkart.
\newblock Learning an animatable detailed {3D} face model from in-the-wild
  images.
\newblock {\em ACM Transactions on Graphics (ToG), Proc. SIGGRAPH},
  40(4):88:1--88:13, 2021.

\bibitem{Goodfellow}
Ian~J. Goodfellow, Jean Pouget-Abadie, Mehdi Mirza, Bing Xu, David
  Warde-Farley, Sherjil Ozair, Aaron Courville, and Yoshua Bengio.
\newblock Generative adversarial nets.
\newblock In {\em Proceedings of the 27th International Conference on Neural
  Information Processing Systems - Volume 2}, NIPS'14, page 2672–2680,
  Cambridge, MA, USA, 2014. MIT Press.

\bibitem{groth2020altering}
Colin Groth, Jan-Philipp Tauscher, Susana Castillo, and Marcus Magnor.
\newblock Altering the conveyed facial emotion through automatic reenactment of
  video portraits.
\newblock In {\em Proceedings of the International Conference on Computer
  Animation and Social Agents ({CASA})}, volume 1300, pages 128--135, 2020.

\bibitem{heusel2017fid}
Martin Heusel, Hubert Ramsauer, Thomas Unterthiner, Bernhard Nessler, and Sepp
  Hochreiter.
\newblock Gans trained by a two time-scale update rule converge to a local nash
  equilibrium.
\newblock In {\em Proceedings of the International Conference on Neural
  Information Processing Systems (NeurIPS)}, 2017.

\bibitem{LSTM}
Sepp Hochreiter and Jürgen Schmidhuber.
\newblock Long short-term memory.
\newblock {\em Neural computation}, 9:1735--80, 12 1997.

\bibitem{pix2pix2017}
Phillip Isola, Jun-Yan Zhu, Tinghui Zhou, and Alexei~A Efros.
\newblock Image-to-image translation with conditional adversarial networks.
\newblock In {\em Proceedings of the IEEE Conference on Computer Vision and
  Pattern Recognition (CVPR)}, 2017.

\bibitem{ji2021audio-driven}
Xinya Ji, Hang Zhou, Kaisiyuan Wang, Wayne Wu, Chen~Change Loy, Xun Cao, and
  Feng Xu.
\newblock Audio-driven emotional video portraits.
\newblock In {\em Proceedings of the IEEE Conference on Computer Vision and
  Pattern Recognition (CVPR)}, 2021.

\bibitem{johnson2021deepfakes}
Deborah~G Johnson and Nicholas Diakopoulos.
\newblock What to do about deepfakes.
\newblock {\em Communications of the ACM}, 64(3):33--35, 2021.

\bibitem{neural_style_preserving_dubbing}
Hyeongwoo Kim, Mohamed Elgharib, Michael Zollh\"{o}fer, Hans-Peter Seidel,
  Thabo Beeler, Christian Richardt, and Christian Theobalt.
\newblock Neural style-preserving visual dubbing.
\newblock {\em ACM Trans. Graph.}, 38(6), Nov. 2019.

\bibitem{DVP}
Hyeongwoo Kim, Pablo Garrido, Ayush Tewari, Weipeng Xu, Justus Thies, Matthias
  Nie{\ss}ner, Patrick P{\'e}rez, Christian Richardt, Michael Zoll{\"o}fer, and
  Christian Theobalt.
\newblock Deep video portraits.
\newblock {\em ACM Transactions on Graphics (TOG)}, 37(4):163, 2018.

\bibitem{kingma2014adam}
Diederik~P Kingma and Jimmy Ba.
\newblock Adam: A method for stochastic optimization.
\newblock {\em arXiv preprint arXiv:1412.6980}, 2014.

\bibitem{kollias2020analysing}
D Kollias, A Schulc, E Hajiyev, and S Zafeiriou.
\newblock Analysing affective behavior in the first abaw 2020 competition.
\newblock In {\em Proceedings of the 15th IEEE International Conference on
  Automatic Face and Gesture Recognition (FG)}, pages 794--800, 2020.

\bibitem{kollias2019face}
Dimitrios Kollias, Viktoriia Sharmanska, and Stefanos Zafeiriou.
\newblock Face behavior a la carte: Expressions, affect and action units in a
  single network.
\newblock {\em arXiv preprint arXiv:1910.11111}, 2019.

\bibitem{kollias2021distribution}
Dimitrios Kollias, Viktoriia Sharmanska, and Stefanos Zafeiriou.
\newblock Distribution matching for heterogeneous multi-task learning: a
  large-scale face study.
\newblock {\em arXiv preprint arXiv:2105.03790}, 2021.

\bibitem{kollias2019deep}
Dimitrios Kollias, Panagiotis Tzirakis, Mihalis~A Nicolaou, Athanasios
  Papaioannou, Guoying Zhao, Bj{\"o}rn Schuller, Irene Kotsia, and Stefanos
  Zafeiriou.
\newblock Deep affect prediction in-the-wild: Aff-wild database and challenge,
  deep architectures, and beyond.
\newblock {\em International Journal of Computer Vision}, pages 1--23, 2019.

\bibitem{kollias2019expression}
Dimitrios Kollias and Stefanos Zafeiriou.
\newblock Expression, affect, action unit recognition: Aff-wild2, multi-task
  learning and arcface.
\newblock {\em arXiv preprint arXiv:1910.04855}, 2019.

\bibitem{kollias2021affect}
Dimitrios Kollias and Stefanos Zafeiriou.
\newblock Affect analysis in-the-wild: Valence-arousal, expressions, action
  units and a unified framework.
\newblock {\em arXiv preprint arXiv:2103.15792}, 2021.

\bibitem{kollias2021analysing}
Dimitrios Kollias and Stefanos Zafeiriou.
\newblock Analysing affective behavior in the second abaw2 competition.
\newblock In {\em Proceedings of the IEEE International Conference on Computer
  Vision (ICCV)}, pages 3652--3660, 2021.

\bibitem{FLAME:SiggraphAsia2017}
Tianye Li, Timo Bolkart, Michael.~J. Black, Hao Li, and Javier Romero.
\newblock Learning a model of facial shape and expression from {4D} scans.
\newblock {\em ACM Transactions on Graphics, (Proc. SIGGRAPH Asia)},
  36(6):194:1--194:17, 2017.

\bibitem{Lindt2019FacialEE}
Alexandra Lindt, Pablo V.~A. Barros, Henrique Siqueira, and Stefan Wermter.
\newblock Facial expression editing with continuous emotion labels.
\newblock {\em Proceedings of the 14th IEEE International Conference on
  Automatic Face \& Gesture Recognition (FG)}, pages 1--8, 2019.

\bibitem{FaceDirector}
Charles Malleson, Jean~Charles Bazin, Oliver Wang, Derek Bradley, Thabo Beeler,
  Adrian Hilton, and Alexander Sorkine-Hornung.
\newblock Facedirector: Continuous control of facial performance in video.
\newblock In {\em Proceedings of the IEEE International Conference on Computer
  Vision (ICCV)}, pages 3979--3987, 2015.

\bibitem{LSGAN}
X. Mao, Q. Li, H. Xie, R.~K. Lau, Z. Wang, and S. Smolley.
\newblock Least squares generative adversarial networks.
\newblock In {\em Proceedings of the IEEE International Conference on Computer
  Vision (ICCV)}, pages 2813--2821, 2017.

\bibitem{mirsky2021creation}
Yisroel Mirsky and Wenke Lee.
\newblock The creation and detection of deepfakes: A survey.
\newblock {\em ACM Computing Surveys (CSUR)}, 54(1):1--41, 2021.

\bibitem{HighResolutionNeuralFaceSwapping}
Jacek Naruniec, Leonhard Helminger, Christopher Schroers, and Romann~M. Weber.
\newblock High-resolution neural face swapping for visual effects.
\newblock {\em Computer Graphics Forum}, 39(4):173--184, 2020.

\bibitem{nirkin2019fsgan}
Yuval Nirkin, Yosi Keller, and Tal Hassner.
\newblock {FSGAN}: Subject agnostic face swapping and reenactment.
\newblock In {\em Proceedings of the IEEE International Conference on Computer
  Vision (ICCV)}, pages 7184--7193, 2019.

\bibitem{poisson_editing}
Patrick P\'{e}rez, Michel Gangnet, and Andrew Blake.
\newblock Poisson image editing.
\newblock {\em ACM Trans. Graph.}, 22(3):313–318, July 2003.

\bibitem{perov2021deepfacelab}
Ivan Perov, Daiheng Gao, Nikolay Chervoniy, Kunlin Liu, Sugasa Marangonda,
  Chris Umé, Mr. Dpfks, Carl~Shift Facenheim, Luis RP, Jian Jiang, Sheng
  Zhang, Pingyu Wu, Bo Zhou, and Weiming Zhang.
\newblock Deepfacelab: Integrated, flexible and extensible face-swapping
  framework, 2021.

\bibitem{DSM21}
Girish~Kumar Solanki and Anastasios Roussos.
\newblock Deep semantic manipulation of facial videos.
\newblock {\em arXiv preprint arXiv:2111.07902}, 2021.

\bibitem{Thies_real_time}
Justus Thies, Michael Zollh\"{o}fer, Matthias Nie\ss{}ner, Levi Valgaerts, Marc
  Stamminger, and Christian Theobalt.
\newblock Real-time expression transfer for facial reenactment.
\newblock {\em ACM Trans. Graph.}, 34(6), Oct. 2015.

\bibitem{face2face}
Justus Thies, Michael Zollh\"{o}fer, Marc Stamminger, Christian Theobalt, and
  Matthias Nie\ss{}ner.
\newblock Face2face: Real-time face capture and reenactment of rgb videos.
\newblock {\em Commun. ACM}, 62(1):96–104, Dec. 2018.

\bibitem{tolosana2020deepfakes}
Ruben Tolosana, Ruben Vera-Rodriguez, Julian Fierrez, Aythami Morales, and
  Javier Ortega-Garcia.
\newblock Deepfakes and beyond: A survey of face manipulation and fake
  detection.
\newblock {\em Information Fusion}, 64:131--148, 2020.

\bibitem{Tripathy_ICface}
Soumya Tripathy, Juho Kannala, and Esa Rahtu.
\newblock Icface: Interpretable and controllable face reenactment using gans.
\newblock In {\em Proceedings of the IEEE/CVF Winter Conference on Applications
  of Computer Vision (WACV)}, 2020.

\bibitem{Tripathy_FACEGAN}
Soumya Tripathy, Juho Kannala, and Esa Rahtu.
\newblock Facegan: Facial attribute controllable reenactment gan.
\newblock In {\em Proceedings of the IEEE/CVF Winter Conference on Applications
  of Computer Vision (WACV)}, 2021.

\bibitem{L1_median}
Yehuda Vardi and Cun-Hui Zhang.
\newblock The multivariate l1-median and associated data depth.
\newblock {\em Proceedings of the National Academy of Sciences of the United
  States of America}, 97:1423--6, 03 2000.

\bibitem{wang2020mead}
Kaisiyuan Wang, Qianyi Wu, Linsen Song, Zhuoqian Yang, Wayne Wu, Chen Qian, Ran
  He, Yu Qiao, and Chen~Change Loy.
\newblock Mead: A large-scale audio-visual dataset for emotional talking-face
  generation.
\newblock In {\em Proceedings of the European Conference on Computer Vision
  (ECCV)}, pages 700--717, 2020.

\bibitem{yadlin2021whose}
Aya Yadlin-Segal and Yael Oppenheim.
\newblock Whose dystopia is it anyway? deepfakes and social media regulation.
\newblock {\em Convergence}, 27(1):36--51, 2021.

\bibitem{yao2021talkinghead}
Xinwei Yao, Ohad Fried, Kayvon Fatahalian, and Maneesh Agrawala.
\newblock Iterative text-based editing of talking-heads using neural
  retargeting.
\newblock {\em ACM Trans. Graph.}, 40(3), Aug. 2021.

\bibitem{yu2021survey}
Peipeng Yu, Zhihua Xia, Jianwei Fei, and Yujiang Lu.
\newblock A survey on deepfake video detection.
\newblock {\em IET Biometrics}, 2021.

\bibitem{zafeiriou2017aff}
Stefanos Zafeiriou, Dimitrios Kollias, Mihalis~A Nicolaou, Athanasios
  Papaioannou, Guoying Zhao, and Irene Kotsia.
\newblock Aff-wild: Valence and arousal ‘in-the-wild’challenge.
\newblock In {\em 2017 IEEE Conference on Computer Vision and Pattern
  Recognition Workshops (CVPRW)}, pages 1980--1987. IEEE, 2017.

\bibitem{zakharov_few-shot_2019}
Egor Zakharov, Aliaksandra Shysheya, Egor Burkov, and Victor Lempitsky.
\newblock Few-{Shot} {Adversarial} {Learning} of {Realistic} {Neural} {Talking}
  {Head} {Models}.
\newblock In {\em Proceedings of the {IEEE} {International} {Conference} on
  {Computer} {Vision} ({ICCV})}, pages 9458--9467, 2019.

\bibitem{MTCNN}
K. Zhang, Z. Zhang, Z. Li, and Y. Qiao.
\newblock Joint face detection and alignment using multitask cascaded
  convolutional networks.
\newblock {\em IEEE Signal Processing Letters}, 23(10):1499--1503, Oct 2016.

\bibitem{Zhou2021EmotionalVC}
Kun Zhou, Berrak Sisman, Rui Liu, and Haizhou Li.
\newblock Emotional voice conversion: Theory, databases and esd.
\newblock {\em ArXiv}, abs/2105.14762, 2021.

\bibitem{CycleGAN}
Jun-Yan {Zhu}, Taesung {Park}, Phillip {Isola}, and Alexei~A. {Efros}.
\newblock Unpaired image-to-image translation using cycle-consistent
  adversarial networks.
\newblock In {\em Proceedings of the IEEE International Conference on Computer
  Vision (ICCV)}, pages 2242--2251, 2017.

\end{thebibliography}
}

\clearpage
\appendix

\twocolumn[{%
\renewcommand\twocolumn[1][]{#1}%
\begin{center}
\textbf{\Large Neural Emotion Director: Speech-preserving semantic control of facial \\ \vspace{3pt} expressions in “in-the-wild” videos}

\vspace{3pt}
\textbf{\Large (Supplementary Material)}
\vspace{30pt}
\end{center}%
}]

\section{Rendering of NMFC and Detailed Shape Images}
\begin{figure*}[h]
\centering
\includegraphics[width=1.0\textwidth]{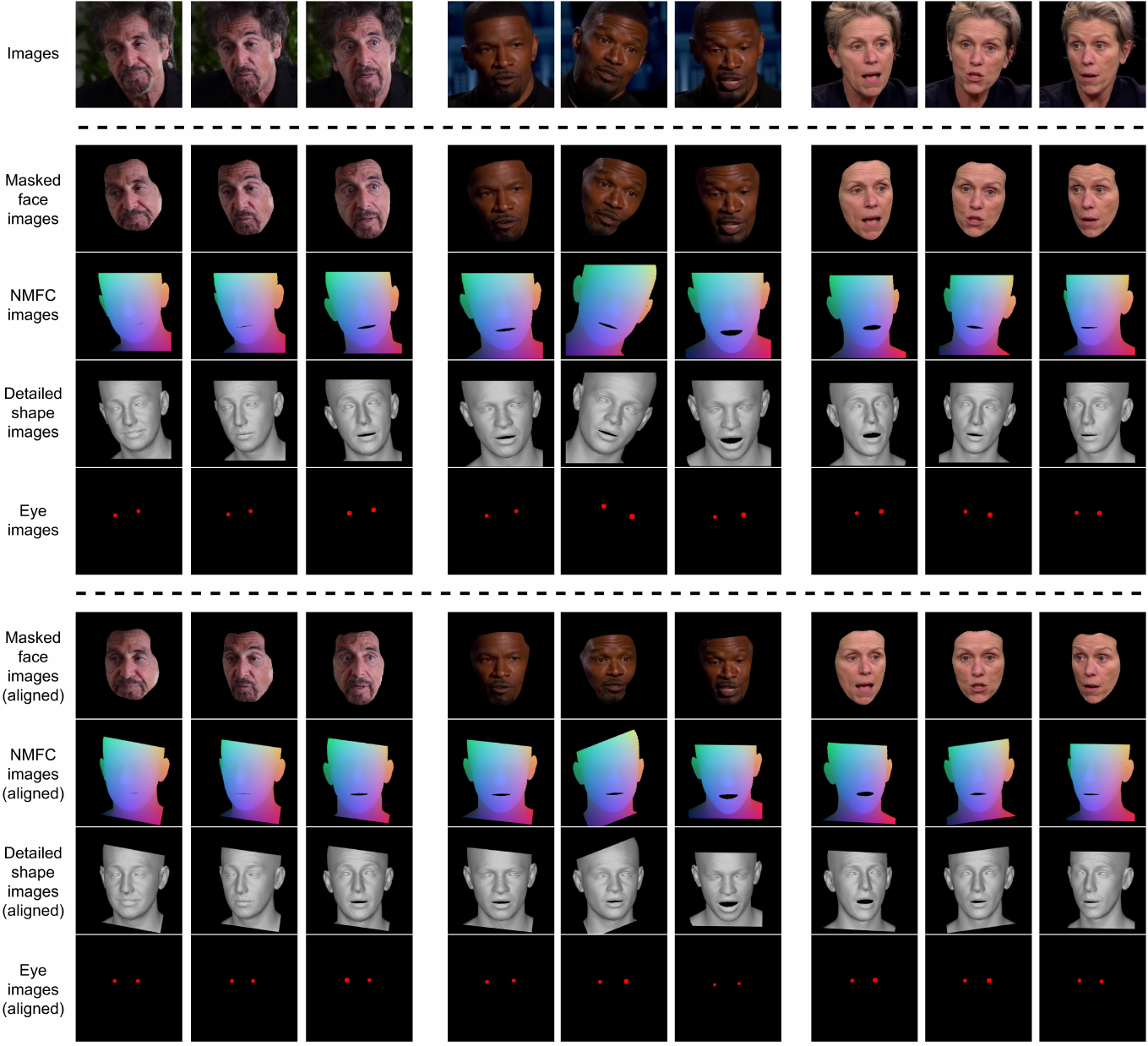}
\caption{Example frames of cropped face images ($1^{st}$ row), their masking after face segmentation ($2^{nd}$ row) as well as their corresponding NMFC, Detailed Shape and Eye images (next 3 rows).  
The last 4 rows visualize the warped versions of the Masked face, NMFC, Detailed Shape and Eye images, as a result of the face alignment that we apply. 
Note that the last 3 rows (aligned versions of NMFC, Detailed Shape and Eye images) correspond to the conditional inputs for our \textit{Neural Face Renderer}. 
Note also that in this visualization, no emotion manipulation has been applied, which corresponds to the self-reenactment scenario used during training of our \textit{Neural Face Renderer}.}
\label{fig:cond_inputs}
\end{figure*}

As mentioned in Sec.~3.3 of the main paper, we follow \cite{head2head++} and map the manipulated 3D face geometry to a convenient representation for our \textit{Neural Face Renderer}. 
In more detail, we render an RGB image of the 3D face mesh under the manipulated expressions and pose of each given frame using the following semantically-meaningful color-coding scheme: Every vertex of the 3D mesh triangulation of 
the adopted FLAME face model~\cite{FLAME:SiggraphAsia2017} is uniquely colored with RGB values that are directly mapped from the XYZ coordinates of the mean face's 3D mesh, after normalizing them to  $[0,1]$, see \eg~$3^{rd}$ row of Fig.~\ref{fig:cond_inputs}. As in \cite{head2head++}, we use the term \textit{NMFC (Normalized Mean Face Coordinate)} images for these rendered RGB images (i.e. $\textbf{NMFC}\in\mathbb{R}^{256\times256\times3}$).

As also  mentioned in the main paper, we use the so-called 
\textbf{detailed shape images} as additional conditional input for our \textit{Neural Face Renderer}. In more detail, DECA~\cite{DECA:Siggraph2021} estimates not only a standard 3D face reconstruction but also a person-specific detail vector $\boldsymbol{\delta}\in\mathbb{R}^{128}$ for each frame, which improves upon previous methods by adding mid-frequency details to the face geometry through a UV displacement map. 
We take advantage of this and deviate from \cite{head2head++} (which is based on older 3D face modelling approaches)  by adding the resultant detailed shape images $\textbf{S}$ as an extra conditional input for the face renderer. To be more precise, the detail vector $\boldsymbol{\delta}$ is used in combination with the manipulated expressions to generate detailed 3D face shapes, which we then render (using conventional 3D graphics as in the case of NMFCs) to create the so-called detailed shape images $\textbf{S}\in\mathbb{R}^{256\times256\times3}$, see \eg~$4^{th}$ row of Fig.~\ref{fig:cond_inputs}. In contrast to NMFCs, for this rendering we use constant gray colouring of the 3D mesh and standard smooth shading, since we want fine geometric details (wrinkles, dimples, etc.) to be clearly visible and help in the conditioning of the neural renderer, in terns of photo-realism of the synthesized face images. 

An illustration example of all types of conditional inputs that we feed our \textit{Neural Face Renderer} with is provided in Fig.~\ref{fig:cond_inputs}. 

\section{Face alignment}
In this section, we provide more details about face alignment, which is one of the main steps in the \textit{3D Face Analysis} module of our pipeline (see  Sec.~3.1 of the main paper).  Face alignment is an essential step in most \textbf{face-swapping} methods (\eg~\cite{perov2021deepfacelab}), where combining footage from 2 different actors in totally different poses requires their faces being brought in correspondence. Although our main aim is to generate the target actor under similar conditions to those seen during training, we found that face alignment is still useful, as it boosts our Neural Face Renderer's generalization ability. In particular, we observed that the renderer 
struggles to produce novel expressions if those are not present in the training footage under the same pose and position. 
Therefore, at each given frame, we use Procrustes analysis to estimate a 2D similarity transformation matrix between the 68 extracted landmarks and the corresponding landmarks of a mean face template. 
The  masked input face images, as well as the NMFC, detailed shape, and eye-gaze images are then warped according to this transformation, see \eg~last 4 rows of Fig.~\ref{fig:cond_inputs}. Furthermore, to avoid jittering artifacts in the aligned images we follow \cite{HighResolutionNeuralFaceSwapping} and average the landmarks extracted from multiple slightly displaced versions of the original face image.

\section{Blending}
Here, we provide more details about Blending, the final step of our \textit{Photo-realistic Synthesis ``in the wild''} module (see Sec.~3.3 of the main paper). 
The seamless composition of the generated face onto the original scene is achieved through multi-band blending~\cite{burt_image_mosaics}. In particular, we construct the Laplacian pyramids of both images and perform blending on each level of the pyramid independently using a softly eroded version of the face mask. The blended image is then obtained by reconstructing it from the final pyramid and is placed in the exact same position of the full frame from where it was cropped, thus, fully reproducing the original video, independently of its spatial resolution.

\section{Training of the Emotion Manipulator: Loss Functions}

In Sec.~3.2.1 of the main paper, we briefly describe the adopted loss functions for training our \textit{Emotion Manipulator}. Here, we expand the presentation providing more details and the relevant mathematical formulas:

The networks of the \textit{Emotion Manipulator} (Translator $G$, Style Encoder $E$, Mapping network $M$ and Discriminator $D$) are updated based on the following loss functions:

\noindent$\bullet$ \textbf{Adversarial loss:} We use LSGAN~\cite{LSGAN} with labels $b$=$c$=1 for real samples and label $a$=0 for fake ones, resulting in the following adversarial objectives for the Discriminator $D$:
\begin{equation}
\mathcal{L}_{adv}^D=\frac{1}{2}\mathbb{E}_{\textbf{s}, y}[(D_y(\textbf{s})-1)^2] + \frac{1}{2}\mathbb{E}_{\textbf{s}, \tilde{y}, \tilde{\textbf{d}}}[D_{\tilde{y}}(G(\textbf{s}, \tilde{\textbf{d}}))^2]
\end{equation}\label{eqn:adv_loss_D}

... and the Translator $G$:
\begin{equation}
\mathcal{L}_{adv}^G=\frac{1}{2}\mathbb{E}_{\textbf{s}, \tilde{y}, \tilde{\textbf{d}}}[(D_{\tilde{y}}(G(\textbf{s}, \tilde{\textbf{d}}))-1)^2]
\label{eqn:adv_loss_G}
\end{equation}

This way the mapping network $M$ learns to output the speaking styles that belong to the emotional domain $\tilde{y}$ and the translator to produce sequences of the target domain that are indistinguishable from the real ones.

\noindent$\bullet$ \textbf{Style reconstruction loss:} As in~\cite{choi2020starganv2}, we make sure the output sequence reflects the given style by using a loss that enforces the style vector of the translated sequence, as extracted by the style encoder $E$, to  match the desired one:
\begin{equation}
\mathcal{L}_{sty}=\mathbb{E}_{\textbf{s}, \tilde{\textbf{d}}}[|| \tilde{\textbf{d}} - E(G(\textbf{s}, \tilde{\textbf{d}}))||_1]
\label{eqn:sty_loss}
\end{equation}
$\bullet$ \textbf{Cycle consistency loss:} We use the cycle consistency loss~\cite{CycleGAN,StarGAN}, which encourages the translator to produce sequences that preserve the content of the input sequence, so that the input sequence can be reconstructed by translating the output sequence back to the original style $\hat{\textbf{d}}=E(\textbf{s})$, as extracted by $E$:
\begin{equation}
\mathcal{L}_{cyc}=\mathbb{E}_{\textbf{s}, \tilde{\textbf{d}}}[|| \textbf{s} - G(G(\textbf{s}, \tilde{\textbf{d}}), \hat{\textbf{d}})||_1] \,.
\label{eqn:cyc_loss}
\end{equation}
$\bullet$ \textbf{Speech-preserving loss:} As observed in~\cite{neural_style_preserving_dubbing}, the cycle consistency loss does not always guarantee that the original mouth motion related to speech is preserved by the translator. To this end, we take advantage of the adopted FLAME face model~\cite{FLAME:SiggraphAsia2017}, which explicitly controls the jaw opening through the $1^{st}$ jaw articulation parameter. Thus, we add an extra constraint to the total objective, that takes into account only this mouth-related parameter, instead of the whole expression vector as in~\cite{neural_style_preserving_dubbing}. We note here that we seek to faithfully alter the conveyed emotion of a sequence and this usually requires increasing or decreasing the mouth opening, \eg to show anger or the neutral emotion respectively. Hence, the jaw opening of the original and the translated sequence have to be highly correlated, but not identical. This leads us to define our speech-preserving loss in terms of the \textbf{Pearson Correlation Coefficient (PCC)} between the original and the translated jaw variable within a sequence:
\begin{equation}
\mathcal{L}_{mouth}=-\mathbb{E}_{\textbf{s}, \tilde{\textbf{d}}}[\rho_{\boldsymbol{\epsilon}_1, \tilde{\boldsymbol{\epsilon}_1}} + \rho_{\tilde{\boldsymbol{\epsilon}}_1, \hat{\boldsymbol{\epsilon}_1}}]
\label{eqn:mouth_loss}
\end{equation}
where $\boldsymbol{\epsilon}_1 \in \mathbb{R}$ denotes the first component of the expression vector (jaw opening) and 
$\rho_{X,Y}$ is the PCC between two variables $X, Y$. 
The negative sign originates from the fact that, ideally, we would like the PCC to be maximized, whereas the mouth loss to be minimized. 
The objective is calculated in a symmetric way, where
$\textbf{s}$ is the original sequence, $\tilde{\textbf{s}}=G(\textbf{s}, \tilde{\textbf{d}})$ the translated and $\hat{\textbf{s}}=G(\tilde{\textbf{s}}, \hat{\textbf{d}})$ the reconstructed one. 
All statistical values are calculated as arithmetic means within the $N$ occurences of a sequence. By maximizing the above loss, we manage to balance our challenging and contradictive goal of altering the emotion without distorting the perceived speech (see Fig.~3 of the main paper).
\\
$\bullet$ \textbf{Overall objectives:} The objective for $G, E$ and $M$ that is minimized during training is the following:
\begin{equation*}
\mathcal{L}^{G,E,M}\!=\!\mathcal{L}_{adv}^G + {\lambda}_{sty} \thinspace \mathcal{L}_{sty} + {\lambda}_{cyc} \thinspace \mathcal{L}_{cyc}  + {\lambda}_{mouth}\thinspace\mathcal{L}_{mouth}
\end{equation*}
whereas the objective for $D$ is: $\mathcal{L}^D$=$\mathcal{L}_{adv}^D$. 

It is worth mentioning that we do not use a diversity sensitive loss as in the original StarGAN v2~\cite{choi2020starganv2}, since we found it to be not necessary. Also, for balancing our objectives we use ${\lambda}_{cyc} = {\lambda}_{sty} = {\lambda}_{mouth} = 1$, since we have experimentally observed that this choice yields high-quality results.

\begin{figure*}[ht]
\centering
\includegraphics[width=1.0\textwidth]{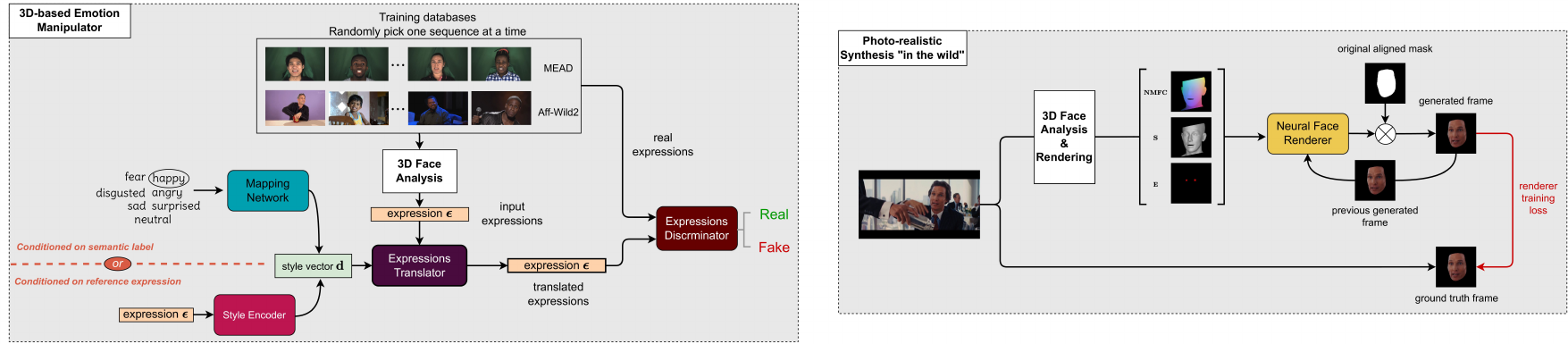}\\
\makebox[0.5\linewidth][c]{\footnotesize{\hspace{3mm}(a) Multi-person training}}\hfill
\makebox[0.5\linewidth][c]{\footnotesize{\hspace{5mm}(b) Person-specific training (fine-tuning)}}\hfill \\
\caption{The 2 trainable components of our \textit{Neural Emotion Director (NED)} are trained separately. \textbf{(a)} The \textit{Emotion Manipulator} is trained on person-agnostic expression data, extracted from 2 large video databases with emotion annotations (Aff-Wild2, MEAD). This trained \textit{Manipulator} can then be used for translating the expressions of any new given actor. \textbf{(b)} The \textit{Neural Face Renderer} is trained independently for each new actor, by fine-tuning the pre-trained meta-renderer on the training footage of the given actor in a self-reenactment fashion.}
\label{fig:arch_train}
\end{figure*}

The \textit{Emotion Manipulator} is trained once (see Fig.~\ref{fig:arch_train} (a)) and can then be run on the fly for altering the expressions of every new actor.

\section{Datasets for Experiments}

In this section, we provide more details about the two datasets used in our experiments: 

\textbf{YouTube Actors dataset:} We collected a small dataset from 6 YouTube videos (having Creative Commons license) that included facial videos of 6 celebrity actors during film scenes, TV shows and interviews under ``in-the-wild'' conditions. Short video clips with duration from 2 to 7 mins that capture the actors during talking and performing were excerpted from these YouTube videos and constituted our dataset. The videos of our YouTube dataset were at 30 fps with a spatial resolution of $1280\times720$ pixels.

\begin{figure*}[t]
\centering
\includegraphics[width=1.0\textwidth]{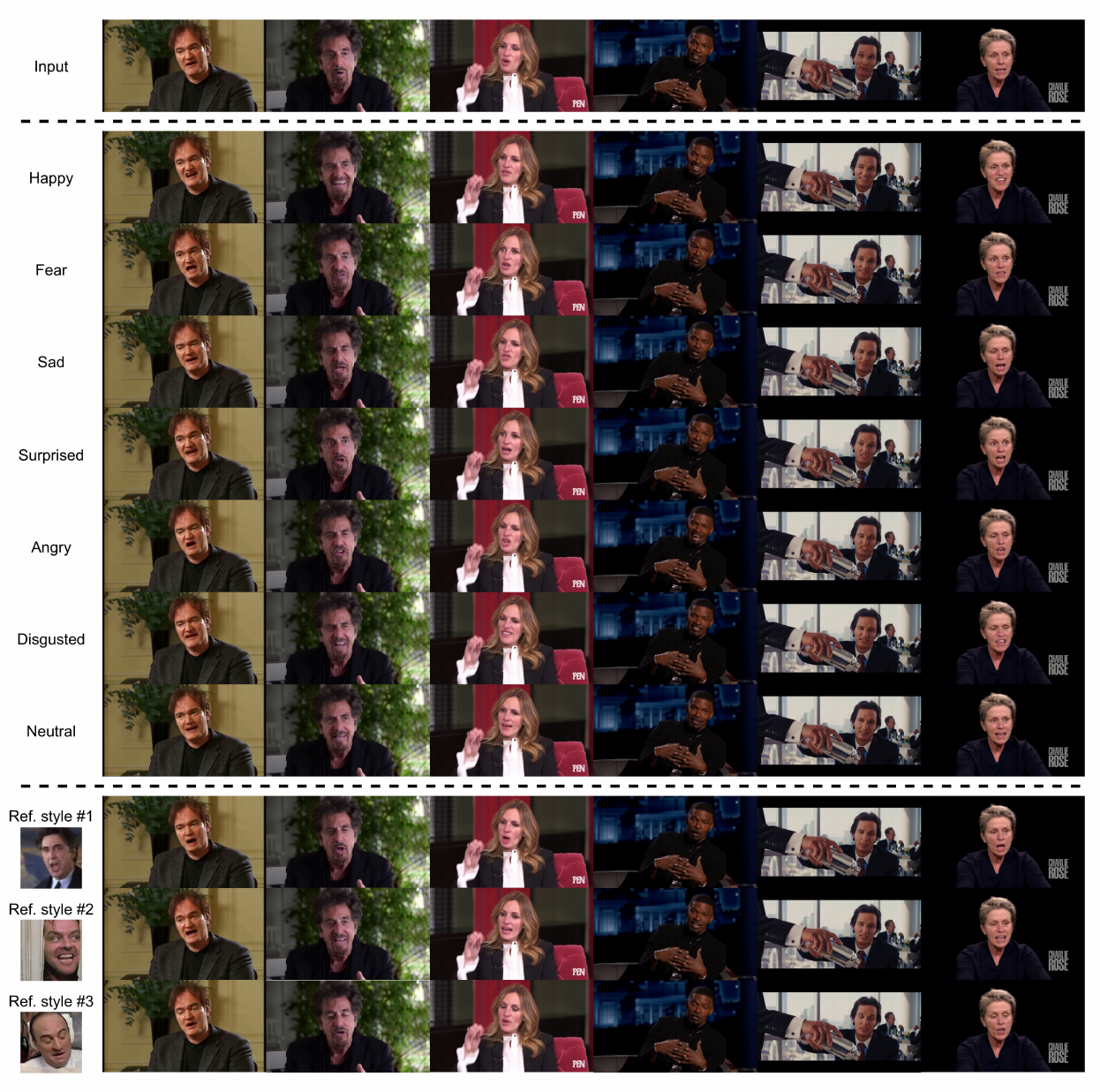}
\caption{Additional visualizations of our results on the YouTube Actors dataset for random frames of the videos. The first row shows the real frames. The next seven rows correspond to expression manipulation based on categorical labels, while the last three rows imitate the expressive style of three exemplar reference clips. Please zoom in for details.}
\label{fig:qual_results}
\end{figure*}

\begin{figure*}[t]
\centering
\includegraphics[width=1.0\textwidth]{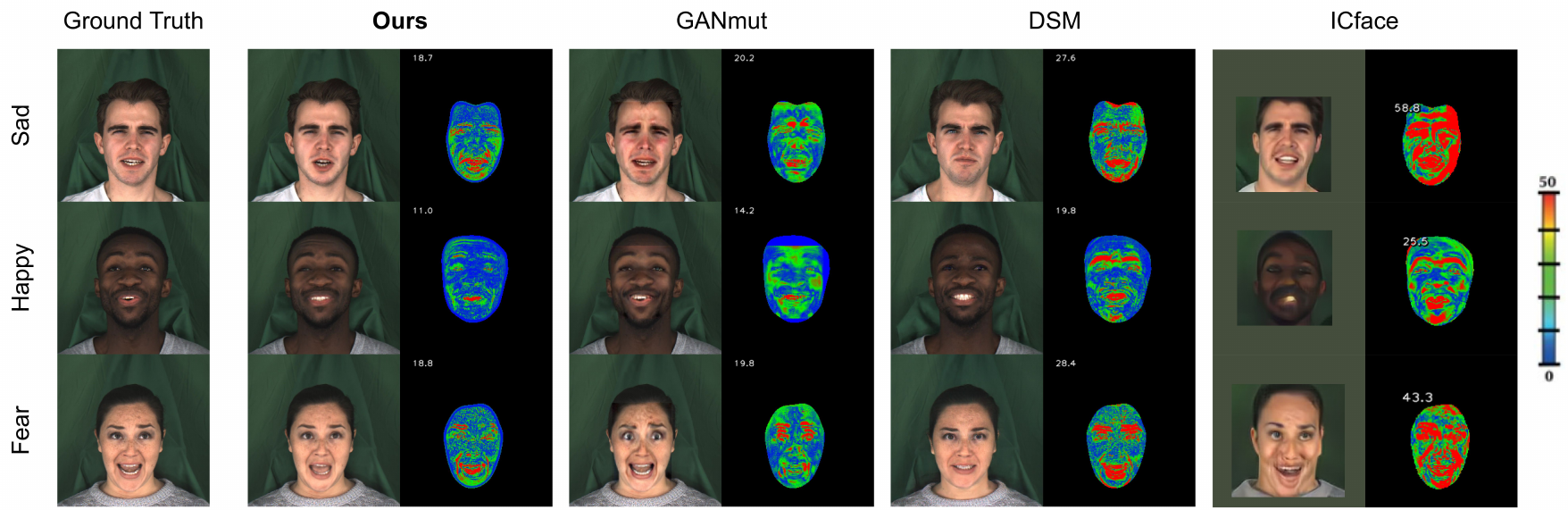}
\caption{Quantitative comparison with state-of-the-art methods in the emotional “self-translation” experiment on the MEAD actors. For each method, the left image is the generated one and the right image is the error between the generated and the input image, visualized as a heat map within the face mask. White numbers inside the heat maps denote the average error for the given frame (Face Average Pixel Distance - FAPD). We observe that, in all cases, our method yields the lowest average errors. Note that the considered range of pixel values is [0,255]. }
\label{fig:heatmaps}
\end{figure*}

\begin{figure*}[h]
\centering
\includegraphics[trim=0 1cm 0.7cm 0.3cm, clip, width=1.0\textwidth]{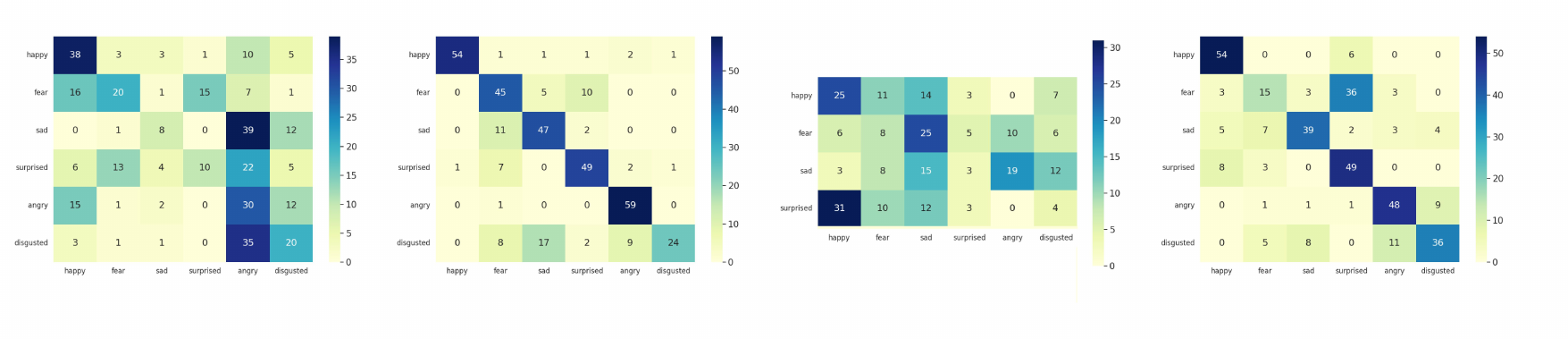}\\
\makebox[0.25\linewidth][c]{\footnotesize{Ours}}\hfill
\makebox[0.25\linewidth][c]{\footnotesize{GANmut}}\hfill
\makebox[0.25\linewidth][c]{\footnotesize{DSM}}\hfill
\makebox[0.25\linewidth][c]{\footnotesize{Ground Truth}}\hfill\\
\caption{Per-method confusion matrices for the classification of emotions regarding the user study on MEAD. Row labelling corresponds to the ground truth annotations, while column labelling to the predicted ones.}
\label{fig:confusion}
\end{figure*}

\textbf{MEAD dataset:} We chose 3 actors from the recent MEAD database~\cite{wang2020mead}. These actors were not included in the training set that we used for our \textit{Emotion Manipulator}. For every actor, we selected 30 videos for each 
of the 6 considered emotions (happy, angry, surprised, fear, sad, disgusted) plus neutral, resulting to a total of 630 videos from MEAD. We note that while this database provides emotional videos in 3 different intensity levels, we use only the ones with the highest intensity per emotion. These videos are at 30 fps, $1920\times1080$ pixels and have an average duration of 4 secs.

Please note that, in both datasets, the selection of the specific actors was done taking into account gender and ethnic group variability. Also, roughly 10\% of each actor's footage was kept as test data for the experiments (and 90\% as train data for the neural face renderer).

\begin{figure}[h]
\centering
\includegraphics[width=0.46\textwidth]{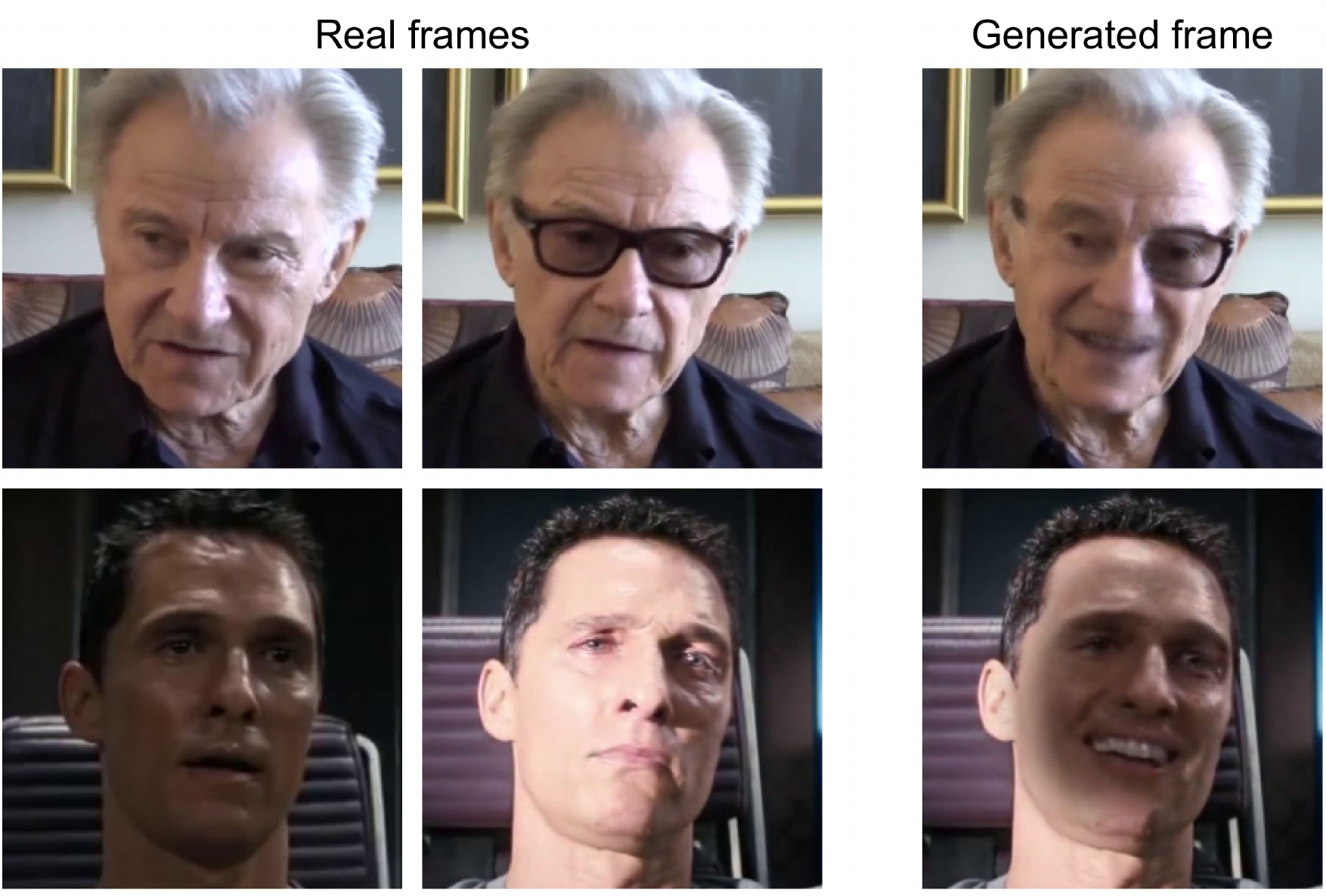}
\caption{Typical failure cases of our \textit{Neural Face Renderer} on challenging videos that include appearance and disappearance of glasses (1$^{st}$ row) or diverse illumination conditions (2$^{nd}$ row). The generator lacks conditional knowledge of which appearance/illumination to produce, thus generating a mixed result. The first 2 frames of each row correspond to real frames seen during training, while the rightmost frame corresponds to the middle frame being translated to \textit{happy}.}
\label{fig:fail_cases}
\end{figure}

\section{Training of the Neural Face Renderer: Meta-renderer}

Our proposed method for increasing the expressive variability of faces generated by our \textit{Emotional Face Renderer} for actors found in YouTube videos, uses the following scheme:
\\
$\bullet$ First, we train a single \textbf{meta-renderer} on a collection of videos consisting of our \textbf{YouTube Actors dataset} (extended with 2 more YouTube actors for greater variability) as well as an extension of our selection from the \textbf{MEAD dataset} that we make for our experiments (made of 2 more MEAD actors, i.e. 5 in total). In this stage, the meta-renderer struggles to disentagle the different actors but learns to transfuse the expressivity of the MEAD actors to the YouTube actors.
\\
$\bullet$ Then we fine-tune our \textbf{meta-renderer} independently for each actor for a few epochs. This step resolves the ``identity confusion" caused by the previous stage, without over-adapting to the given actor's expressions.
\\
We found that fine-tuning the \textbf{meta-renderer} on a new actor that was not used in the the first stage has similar effects. This is especially important because it means that one does not have to repeat the lengthy pre-training each time he/she wants to apply the method to a new actor. \textbf{In other words, for manipulating a new video with a never-before-seen face, the only model that needs to be trained is a new person-specific neural face renderer (generator and discriminator) for synthesizing the new face}. For this, we can simply use the input video as training footage to fine-tune the pre-trained meta-renderer (see Fig.~\ref{fig:arch_train} (b)). Nevertheless, performance slighly increases when the actor's footage is used in the training of the \textbf{meta-renderer}. Therefore, for obtaining the results reported in the main paper all our actors were included in the training of the \textbf{meta-renderer}. Overall, for multi-person pre-training (meta-renderer) we used $\sim$200K frames (60\% from MEAD, 40\% from YouTube), while for person-specific fine-tuning we typically use $\sim$ 5-13K frames.

\section{Additional Qualitative Results}
We provide additional qualitative results of our method in the form of static frames for the actors of our YouTube dataset in Fig.~\ref{fig:qual_results}. Our method translates the expressions of the actors to any of the 6 basic emotions plus neutral or to any given reference style, regardless of the emotion of the input video. We observe that we achieve highly-realistic results. 

In addition, Fig.~\ref{fig:heatmaps} visualizes some random cases from the quantitative comparison of Sec.~4.1 of the main paper. In particular, we visualize the pixel distance between the ``self-translated" frame (i.e. translated to the same emotion that it is labelled as) and ground truth frame as a heat map in the face area. As also reported in the main paper, we observe that our method performs better in preserving the original emotion without distorting the characteristics of the specific identity, which results in lower FAPD values than all the other methods.

\section{Detailed scores for the user study on MEAD database}
\begin{table*}[h!]
\setlength{\tabcolsep}{2pt}
\scriptsize
\centering
\begin{tabular}{l|cccccc|cccccc|cccccc|cccccc|cccc}
\toprule
{} & \multicolumn{24}{c|}{Realism} & \multicolumn{4}{c}{Accuracy} \\
\midrule
{} & \multicolumn{6}{c|}{Ours} & \multicolumn{6}{c|}{GANmut} & \multicolumn{6}{c|}{DSM} & \multicolumn{6}{c|}{Ground Truth} & \multirow{ 2}{*}{Ours} & \multirow{ 2}{*}{GANmut} & \multirow{ 2}{*}{DSM} & \multirow{ 2}{*}{Ground Truth} \\  
{} &    1 &  2 &  3 &  4 & 5 & real &      1 &  2 &  3 &  4 & 5 & real &      1 &  2 &  3 & 4 & 5 & real &            1 &  2 &  3 &  4 &  5 & real &  \multicolumn{4}{c}{} \\
\midrule
happy     &   15 & 17 & \textbf{18} & 10 & 0 &   17\% &     \textbf{39} &  8 & 11 &  1 & 1 &    3\% &     \textbf{26} & 23 &  6 & 2 & 3 &    8\% &            0 &  1 & 11 & 15 & \textbf{33} &   80\% &   63\% &     90\% &     42\% &           90\% \\
fear      &    5 & 10 & \textbf{26} & 15 & 4 &   32\% &     \textbf{34} & 14 &  8 &  2 & 2 &    7\% &     19 & \textbf{27} &  8 & 5 & 1 &   10\% &            2 &  4 & 14 & 20 & \textbf{20} &   67\% &   33\% &     75\% &     13\% &           25\% \\
sad       &    4 & 15 & \textbf{23} & 15 & 3 &   30\% &      9 & \textbf{22} & 18 &  7 & 4 &   18\% &     \textbf{20} & 19 & 14 & 7 & 0 &   12\% &            2 & 11 & 14 & 15 & \textbf{18} &   55\% &   13\% &     78\% &     25\% &           65\% \\
surprised &    7 & \textbf{21} & 19 & 12 & 1 &   22\% &     \textbf{35} & 13 &  7 &  4 & 1 &    8\% &     19 & \textbf{25} & 12 & 3 & 1 &    7\% &            0 &  1 & 10 & 20 & \textbf{29} &   82\% &   17\% &     82\% &      5\% &           82\% \\
angry     &    3 & 18 & \textbf{24} &  9 & 6 &   25\% &     \textbf{38} & 11 &  5 &  3 & 3 &   10\% &      - &  - &  - & - & - &    - &            0 &  2 & 11 & \textbf{29} & 18 &   78\% &   50\% &     98\% &      - &           80\% \\
disgusted &    1 & 17 & \textbf{18} & 20 & 4 &   40\% &     12 & \textbf{25} & 11 & 11 & 1 &   20\% &      - &  - &  - & - & - &    - &            2 &  3 & 15 & \textbf{21} & 19 &   67\% &   33\% &     40\% &      - &           60\% \\ \midrule
avg.   &    6 & 16 & \textbf{21} & 14 & 4 &   28\% &     \textbf{28} & 16 & 10 &  5 & 2 &   11\% &     21 & \textbf{24} & 10 & 4 & 2 &    9\% &            2 &  4 & 12 & 20 & \textbf{23} &   71\% &   35\% &     77\% &     21\% &           67\% \\
\bottomrule
\end{tabular}
 \caption{Detailed realism ratings and emotion classification accuracy of the user study on MEAD. Columns 1-5 show the number of times that users gave this realism rating. The column ``real" shows the percentage of users that rated the videos with 4 or 5. Bold values denote the most frequent user realism rating for each method and emotion.}
 \label{tab:user_study_2}
\end{table*}

As mentioned in Sec.~4.2 of the main paper, our second user study asked participants to both evaluate the realism and recognize the emotion of the videos shown to them (on MEAD actors). In Tab.~\ref{tab:user_study_2} we provide a more extended presentation of the relevant results, by providing detailed realism scores for each of the five points of the adopted Likert scale. As can be seen, videos generated with our method were most frequently rated with 3, while the other methods achieved 1 or 2 as the most frequent rating. This holds true even for each basic emotion individually. In particular, \textit{surprised} is the only category where our most preferred rating dropped to 2, whereas for the other methods, only 2 of the considered categories yielded a most frequent rating higher than 1 (\textit{sad} and \textit{disgusted} for GANmut~\cite{sdapolito2021GANmut}, \textit{fear} and \textit{surprised} for DSM~\cite{DSM21}).

Finally, we further elaborate on the emotion recognition results of Tab.~\ref{tab:user_study_2} through confusion matrices in Fig.~\ref{fig:confusion}. As can be seen, \textit{sad} and \textit{surprised} are mostly recognized as \textit{angry} for our method, while the misclassifications of the DSM results seem more mixed and unclear. Moreover, users struggle to separate \textit{fear} and \textit{surprised} in real videos, whereas this is successfully achieved for GANmut, confirming that GANmut emotions seem more ‘stereotypical’ than the real ones produced by the actors.

\section{Details about Running the Methods}

For GANmut~\cite{sdapolito2021GANmut} and ICface~\cite{Tripathy_ICface}, we use the codes made publicly available by the authors, whereas for DSM~\cite{DSM21} we use the code that the authors provided us with. In all cases, we use the default parameters specified by the authors.

All methods were run on a machine with 4 NVIDIA RTX 2080 GPU. For training our \textit{Neural Face Renderer}, we used Adam optimizer~\cite{kingma2014adam} with learning rate of 2e-4 for the meta-renderer or 4e-5 for the person-specific fine-tuning, $\beta_1$=0.5, $\beta_2$=0.999, and a batch size of 2. The \textit{3D-based Emotion Manipulator} is trained with the Adam optimizer as well, using a learning rate of 1e-4, $\beta_1$=0, $\beta_2$=0.99 and a batch size of 64. The meta-renderer takes approximately 25 hours to complete training (15 epochs) using 4 GPUs, whereas fine-tuning an actor-specific renderer with 2 GPUs for 20 epochs takes an average of 17 hours. The average end-to-end inference time corresponds to $\sim$4 frames per second.

\section{Neural Face Renderer failure cases}
We note here that our face renderer assumes that the same foreground face is captured in a consistent scene throughout the whole duration of the training video. If this assumption does not hold (\eg~multiple faces appear in the foreground or the video contains diverse scenes with substantially different illumination conditions or actor's appearances in terms of makeup, facial hair, glasses etc), then our method will yield unrealistic results, like the mixed faces shown in Fig.~\ref{fig:fail_cases}. 

\end{document}